%% file: main.tex
\documentclass[10pt,twocolumn,letterpaper]{article}

\usepackage[pagenumbers]{cvpr}              

\usepackage[accsupp]{axessibility} 
\usepackage{graphicx}
\usepackage{amsmath}
\usepackage{amssymb}
\usepackage{booktabs}
\usepackage{multirow}
\usepackage{makecell}
\usepackage[normalem]{ulem}
\useunder{\uline}{\ul}{}

\input{preamble}

\definecolor{cvprblue}{rgb}{0.21,0.49,0.74}
\usepackage[pagebackref,breaklinks,colorlinks,citecolor=cvprblue]{hyperref}

\newcommand{\midsepremove}{\aboverulesep = 0.3025mm \belowrulesep = 0.492mm}
\newcommand{\midsepdefault}{\aboverulesep = 0.605mm \belowrulesep = 0.984mm}
\newcommand{\argmin}{\mathop{\mathrm{argmin}}}

\newcommand{\norm}[1]{\left\|#1\right\|}

\def\paperID{XXXX} 
\def\confName{CVPR}
\def\confYear{2024}

\title{StraightPCF: Straight Point Cloud Filtering}

\author{Dasith de Silva Edirimuni$^1$,~Xuequan Lu$^2$\thanks{Corresponding author: X. Lu, supported by 
fund 3.2501.11.47.
}~,~Gang Li$^1$,~Lei Wei$^1$,~Antonio Robles-Kelly$^1$,~Hongdong Li$^3$ \\
$^1$Deakin University, $^2$La Trobe University, $^3$Australian National University \\
\tt\small{\{dtdesilva, gang.li, lei.wei, antonio.robles-kelly\}@deakin.edu.au,}
\\
\tt\small{b.lu@latrobe.edu.au, hongdong.li@anu.edu.au}
}

\begin{document}
\maketitle

\begin{abstract}
Point cloud filtering is a fundamental 3D vision task, which aims to remove noise while recovering the underlying clean surfaces. State-of-the-art methods remove noise by moving noisy points along stochastic trajectories to the clean surfaces. These methods often require regularization within the training objective and/or during post-processing, to ensure fidelity. In this paper, we introduce StraightPCF, a new deep learning based method for point cloud filtering. It works by moving noisy points along straight paths, thus reducing discretization errors while ensuring faster convergence to the clean surfaces. We model noisy patches as intermediate states between high noise patch variants and their clean counterparts, and design the VelocityModule to infer a constant flow velocity from the former to the latter. This constant flow leads to straight filtering trajectories. In addition, we introduce a DistanceModule that scales the straight trajectory using an estimated distance scalar to attain convergence near the clean surface. Our network is lightweight and only has $\sim530K$ parameters, being 17\% of IterativePFN (a most recent point cloud filtering network). Extensive experiments on both synthetic and real-world data show our method achieves state-of-the-art results. Our method also demonstrates nice distributions of filtered points without the need for regularization. The implementation code can be found at: \url{https://github.com/ddsediri/StraightPCF}. 
\end{abstract}

\section{Introduction}
\label{sec:intro}
In recent years, point clouds have become increasingly popular as the representation-of-choice for storing and manipulating 3D data, 
with numerous applications in both computer vision~\cite{Luo-Pillar-Motion, Geiger-KITTI, Serna-Rue-Madame, Liao-Kitti360} and geometry modelling
~\cite{Kim-3D-Printing, Urech-Urban-Planning, Luo-Pillar-Motion}. Point clouds are unordered sets of 3D coordinates which typically represent object surfaces and are 
captured using 3D sensors such as depth and Lidar devices. However, noisy artifacts may appear in point clouds as a result of sensor limitations and environmental factors. Removing this noise, known as filtering or denoising, is a fundamental 3D vision task. 
Filtering methods 
are broadly categorized into two 
groups: 1) conventional methods involving traditional optimization and 2) deep learning based methods. Conventional methods can further be subdivided into normal based methods which require surface normal information to reliably filter point clouds~\cite{Alexa--MLS-PSS, APSS-Guennebaud, Digne-Bilateral, Lu-Low-Rank, Mattei-MRPCA} and point based methods which directly filter point clouds~\cite{Lipman-LOP, Huang-WLOP, Preiner-CLOP}. While normal based methods are limited by the accuracy of 
normals, point based methods suffer from a loss of geometric details and still exhibit sensitivity to noise. Recently, many deep learning approaches have been proposed to overcome these shortcomings.

\begin{figure}[!tp]
\centering
\includegraphics[width=\linewidth]{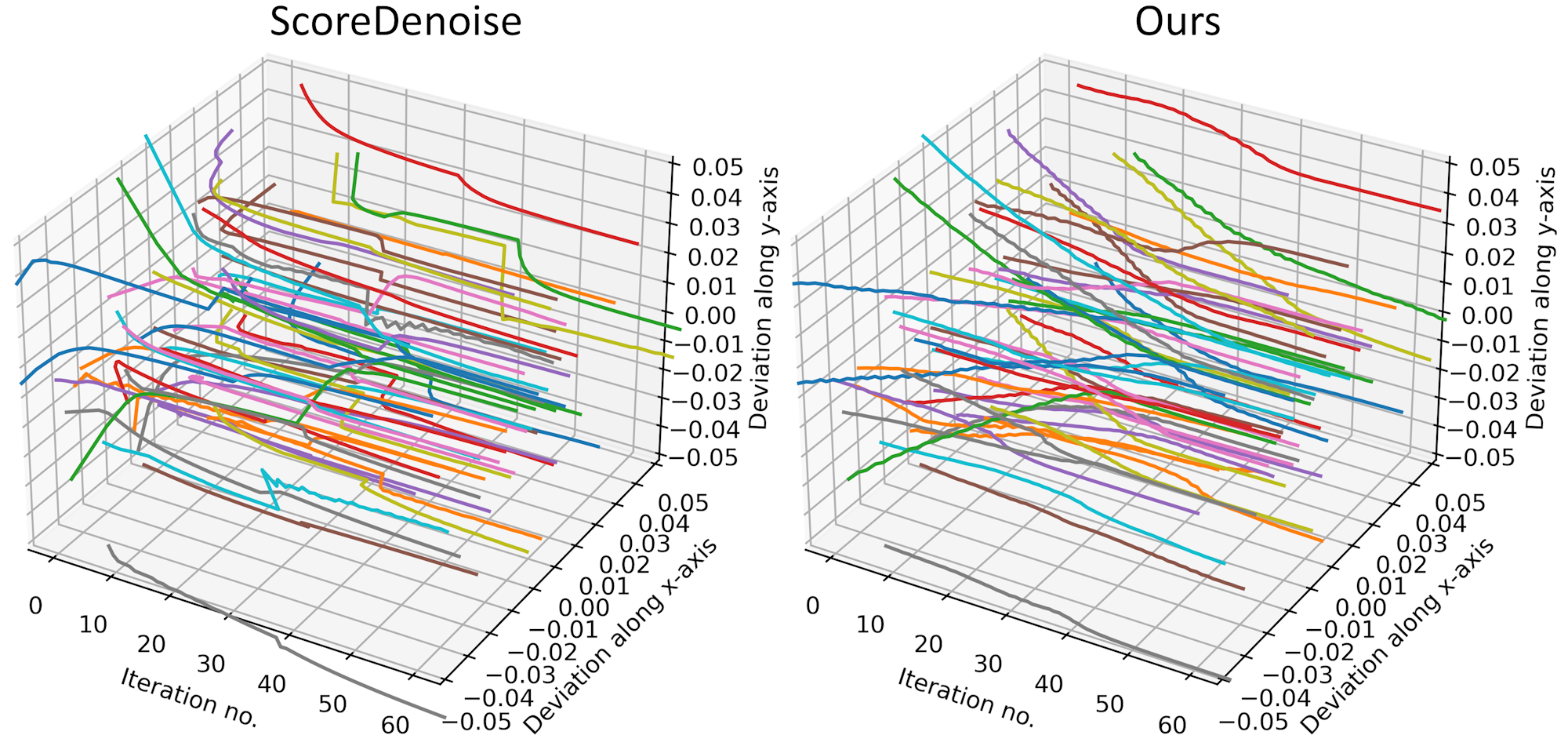}
\caption{Filtered trajectories for the Isocahedron shape at 50K resolution and noise scale $\sigma=3\%$. Our StraightPCF filters points along much straighter paths, compared to ScoreDenoise~\cite{Luo-Score-Based-Denoising}.}
\label{fig:straight-flows}
\end{figure}

Deep learning methods can be divided into 1) \textbf{resampling}, 2) \textbf{displacement} and 3) \textbf{probability} based methods. Resampling-based methods~\cite{Luo-DMRDenoise} show the least fidelity when recovering the underlying noise-free surfaces as their downsampling procedure results in the loss of crucial geometric information. By contrast, displacement and probability based methods show greater promise as they model the filtering objective as a reverse Markov process, which can be iteratively applied on the input to progressively remove noise. Early displacement based methods, such as PointCleanNet~\cite{Rakotosaona-PCN, Zhang-Pointfilter, Edirimuni-Contrastive-Joint-Learning}, employ large networks ($>$1M parameters) that consume a large patch of points to filter a single, central point. This is time-consuming and resource intensive. More recently, IterativePFN~\cite{Edirimuni-IterativePFN} models the iterative filtering process internally using multiple \textit{IterationModules} and filters all patch points simultaneously. However, incorporating multiple IterationModules results in a large network ($>$3.2M parameters) which requires a large amount of GPU memory to process. Probabilistic score based methods such as ScoreDenoise~\cite{Luo-Score-Based-Denoising, Chen-DeepPSR} offer a more lightweight network but require a high number of iterations to recover the clean surface. At higher noise levels, they converge to surfaces that retain noticeable amounts of noise. DeepPSR~\cite{Chen-DeepPSR} utilizes the score module of~\cite{Luo-Score-Based-Denoising} but performs an additional Graph Laplacian Regularization~\cite{Hu-GLR} step on intermediate point clouds to obtain a better point distribution. More importantly, the iterative filtering objectives of current displacement and probabilistic methods move noisy points to the clean surface via stochastic trajectories 
and are error prone as incorrectly inferred displacements may result in significant changes to the distribution of filtered points.

\begin{figure*}[!tp]
    \centering
    \includegraphics[width=0.9\linewidth]{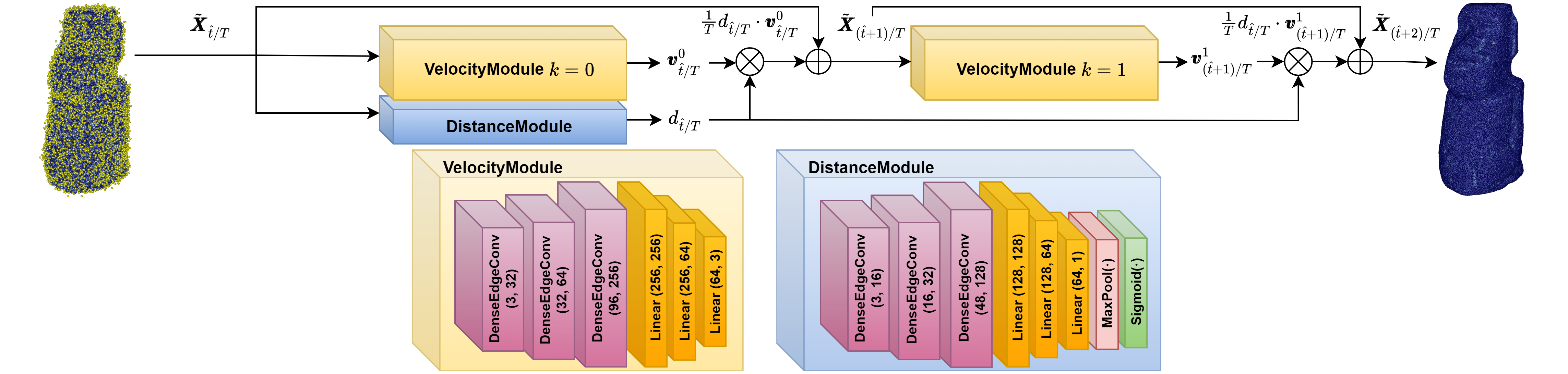}
    \caption{Our StraightPCF network. It involves a coupled VelocityModule stack that infers a constant flow velocity $\pmb{v}^k_\theta$ for patch states $\Tilde{\pmb{X}}_{(\hat{t}+k)/T}$. To ensure filtered points converge to the surface, the DistanceModule infers a distance scalar $d_\phi$ that scales the velocity.}
    \label{fig:straightpcf-network}
\end{figure*}

We propose StraightPCF, which moves noisy points along straight filtering trajectories towards the clean surfaces, as illustrated in Fig.~\ref{fig:straight-flows}. StraightPCF, illustrated in Fig.~\ref{fig:straightpcf-network}, is a lightweight network ($\sim 530K$ parameters, 17\% of IterativePFN~\cite{Edirimuni-IterativePFN}) which filters all patch points simultaneously. Our technical contributions are as follows. 
\begin{itemize}
    \item We introduce the patch-wise \textbf{VelocityModule} that infers constant, straight flows to filter point cloud patches. The VelocityModule elegantly recovers the underlying clean surfaces with nice point distributions.
    \item To improve straightness of flows, we propose a novel straightening mechanism consisting of coupled VelocityModules. This \textbf{coupled VelocityModule stack} infers straighter filtering trajectories, leading to better results.
    \item Constant flows may lead to filtered points overshooting the clean surface. Therefore, we design the \textbf{DistanceModule} to infer a distance scalar that provides a magnitude to scale the flow velocity. Our architecture is the first to decompose filtering into a dual objective of inferring a vector field of flow velocities and a distance scalar.
\end{itemize}

\section{Related Work}
\label{sec:review}
\noindent\textbf{Conventional filtering.}
Early conventional methods were inspired by the Moving Least Squares (MLS) method of Levin~\cite{MLS-Levin} and require normal information for filtering. Notably, Alexa \emph{et al.}~\cite{Alexa--MLS-PSS} employed MLS optimization in recovering denoised surfaces from noisy point sets. The Implicit MLS (IMLS) method of Adamson and Alexa~\cite{IMLS-Adamson} further extended this approach to point-sampled cell complexes which allow for a well defined local geometry. The Algebraic Point Set Surface (APSS) method of Guennebaud and Gross~\cite{APSS-Guennebaud} applied MLS optimization for the purpose of fitting algebraic spheres to recover surfaces while being robust to point set density and underlying curvature. Furthermore, Digne proposed filtering the height map associated to a point set by considering handcrafted features that encode height variations around each point~\cite{Digne-Similarity}. Digne and de Franchis designed a weighted projection scheme, that moves points to their filtered positions~\cite{Digne-Bilateral}, based on the mesh bilateral filtering method of Fleishman, Drori and Cohen-Or~\cite{Fleishman-Bilateral}.
Other normal based methods include 
the Moving Robust Principal Component Analysis (MRPCA)~\cite{Mattei-MRPCA} of Mattei and Castrodad, the Graph Laplacian Regularization (GLR)~\cite{Hu-GLR} technique of Hu \emph{et al.} and the Low Rank Matrix Approximation~\cite{Lu-Low-Rank} of Lu \emph{et al.} The main drawback of such methods is susceptibility to noise, during both the normal estimation and filtering steps. 

By contrast, point based methods employ only point information for filtering. Cazals and Pouget~\cite{Cazals-Jet} proposed a $N$-dimensional polynomial surface fitting method that can be used to filter points. Meanwhile, the Locally Optimal Projection (LOP)~\cite{Lipman-LOP} method of Lipman \emph{et al.} downsampled and regularized noisy point clouds. It was extended by Huang \emph{et al.} and Preiner \emph{et al.} who developed Weighted-LOP (WLOP)~\cite{Huang-WLOP} and Continuous-LOP (CLOP)~\cite{Preiner-CLOP}, respectively. However, these methods do not effectively recover geometric details due to their downsampling step.

\noindent\textbf{Deep learning based filtering.}
While conventional methods rely on handcrafted features, convolutional neural network architectures have provided great improvements to feature generation. PointProNets~\cite{Roveri-PointProNets} by Roveri \emph{et al.} and Deep Feature Preserving (DFP)~\cite{Lu-Deep-Feature-Preserving} by Lu \emph{et al.} projected points onto 2D height maps before processing them using CNNs. PointNet~\cite{Qi-PointNet}, introduced by Qi \emph{et al.}, set the precedent for direct point set convolution and was later improved by PointNet++~\cite{Qi-PointNet++}. Meanwhile, DGCNN~\cite{Wang-DGCNN} of Wang \emph{et al.} reformulated point set convolution as a graph convolution task. PointCleanNet (PCN)~\cite{Rakotosaona-PCN} was one of the first methods to adopt the PointNet architecture and inferred the filtered displacement of a single central point by considering its neighborhood patch. Pointfilter~\cite{Zhang-Pointfilter} of Zhang \emph{et al.} furthered this line of research using a bilateral filtering-inspired loss.

Pistilli \emph{et al.} introduced the first graph convolution-based mechanism, named GPDNet~\cite{Pistilli-GPDNet}, while Luo and Hu proposed the DGCNN based DMRDenoise~\cite{Luo-DMRDenoise}. DMRDenoise filtered points by downsampling noisy inputs and upsampling these less noisy surfaces. Luo and Hu also proposed ScoreDenoise~\cite{Luo-Score-Based-Denoising} where they formally expressed the filtering objective as the backward Langevin equation that iteratively removes noise using the inferred gradient-log of the probability distribution, i.e., the score, for point positions $\pmb{x}$. This was extended by Chen \emph{et al.} in their DeepPSR~\cite{Chen-DeepPSR} which employs an additional graph laplacian regularization post-processing step. Mao \emph{et al.} introduced the normalizing flows based filtering method PDFlow~\cite{Mao-PDFlow} that disentangles noise from the underlying clean representation at higher dimensions. The RePCDNet~\cite{Chen-RePCD} method of Chen \emph{et al.} sought to model iterative filtering via a recurrent neural network. By contrast, de Silva Edirimuni \emph{et al.} proposed the graph convolution based IterativePFN~\cite{Edirimuni-IterativePFN} that models iterative filtering using individual IterationModules. The joint filtering and normal estimation method CFilter~\cite{Edirimuni-Contrastive-Joint-Learning}, developed by de Silva Edirimuni \emph{et al.}, exploited normals to improve filtered point positions. Ma \emph{et al.} introduced learning implicit signed distance functions, by displacing noisy query points back to the surface along normals where the surface corresponds to the zero-level set of the function~\cite{Ma-Neural-Pull}. This has been exploited for normal estimation~\cite{Li-NeuralGF, Li-SHS-Net}.

Score and displacement based methods are inspired by diffusive processes~\cite{Song-Score-Matching, Luo-Diffusive} and their filtering objectives result in stochastic trajectories. Recent work by Liu, Gong and Liu focuses on Reflow~\cite{Liu-RectifiedFlow}, which explored the optimal transport problem of identifying straight paths given two samples from different distributions. The work of Wu \emph{et al.} further extended this Reflow mechanism to point cloud generation, which attempts to generate point cloud shapes given initial samples from the normal distribution~\cite{Wu-PointStraightFlow}.

\section{Problem Statement and Motivation}
\label{sec:motivation}
Resampling methods demonstrate an inability to recover clean surfaces with high accuracy, unlike probabilistic and displacement based methods. A probabilistic score based method such as ScoreDenoise~\cite{Luo-Score-Based-Denoising} and a displacement based method such as IterativePFN~\cite{Edirimuni-IterativePFN} share similar filtering objectives that may be expressed by,
\begin{equation}
\label{eq:pcn-position-update}
    \Tilde{\pmb{x}}^i_t = \Tilde{\pmb{x}}^i_{t-1} + F_\theta(\Tilde{\pmb{x}}^i_{t-1}), t = 1,\ldots,T
\end{equation}
where $\Tilde{\pmb{x}}^i_t$ is the $i$-th filtered point at time $t$ and $F_\theta$ represents the network with parameters $\theta$. The main difference between ScoreDenoise and IterativePFN lies in their training objectives, $\mathcal{L}_S$ and $\mathcal{L}_I$, respectively. For a noisy point $\pmb{x}^i \in \pmb{X}$ where $\pmb{X}$ is the initial noisy patch, we have:
\begin{equation}
    \label{eq:sd-loss}
    \mathcal{L}^i_S = \mathbb{E}_{\pmb{x}\sim kNN(\pmb{x}^i, \pmb{X})}\left[\norm{s(\pmb{x})-\mathcal{S}^i(\pmb{x})}^2_2\right], \\
\end{equation}
\begin{equation}
    \label{eq:ipfn-loss}
    \mathcal{L}^i_I = \sum^T_{t=0}\norm{\pmb{d}^i_t - \left(NN(\pmb{x}^i_{t-1}, \pmb{Y}_t) - \pmb{x}^i_{t-1}\right)}^2_2.
\end{equation}

\begin{figure}[!tp]
    \centering
    \includegraphics[width=0.98\linewidth]{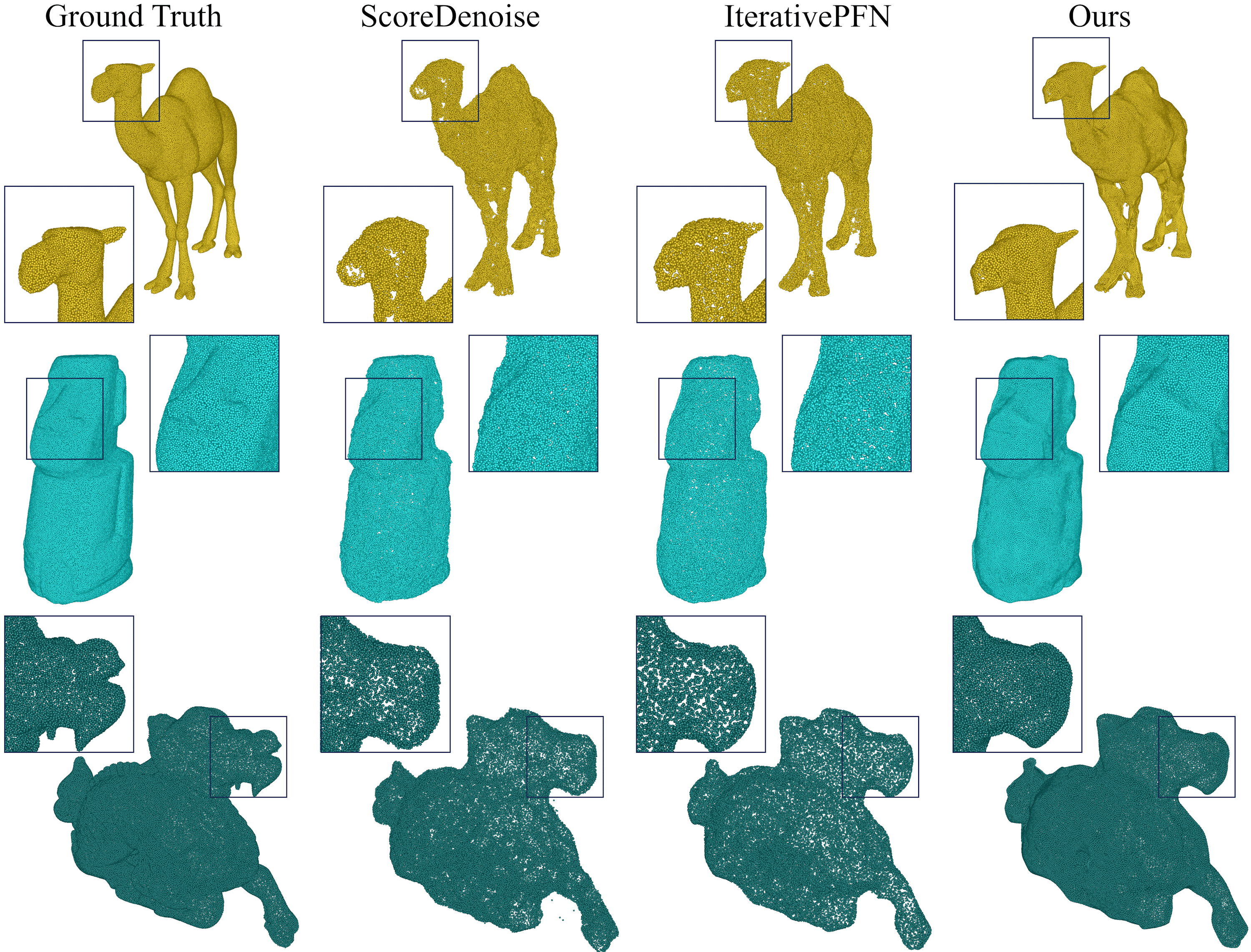}
    \caption{Our StraightPCF is able to recover a better distribution of filtered points, even at very high noise scales ($\sigma=3\%$ and 50K resolution) unseen during network training.}
    \label{fig:point-distribution}
\end{figure}

ScoreDenoise is trained to predict a score $\mathcal{S}^i(\pmb{x}) = \text{Score}(\pmb{x} -\pmb{x}^i|\pmb{h}^i)$ for $\pmb{x}$ sampled from the k-nearest neighbors of $\pmb{x}^i$. It is conditioned on the latent feature $\pmb{h}^i$ of $\pmb{x}^i$ and corresponds to the gradient-log of the noise-convolved probability distribution for point position $\pmb{x}$. The ground truth target $s(\pmb{x}) = NN(\pmb{x}, \pmb{Y}) - \pmb{x}$, where $\pmb{Y}$ is the clean patch. Ergo, $s(\pmb{x})$ is the displacement from $\pmb{x}$ to its nearest neighbor in $\pmb{Y}$. Once this training objective has been optimized, score based methods infer aggregate scores, at each iteration $t$, such that $F_\theta(\pmb{x}^i_{t-1}) = \alpha\mathcal{E}^i(\pmb{x}^i_{t-1}) = (\alpha/K)\sum_{\pmb{x}^j_{t-1} \in kNN(\pmb{x}^i_{t-1})}\mathcal{S}^j(\pmb{x}^i_{t-1})$ where $\alpha$ is the step discretization parameter. This score shifts points towards the clean surface along a stochastic filtered trajectory that is sensitive to the choice of $\alpha$, which must be kept relatively small ($\sim0.2$). By contrast, IterativePFN internally models the iterative filtering objective and incorporates it within their training objective. They directly infer displacements $F_\theta(\Tilde{\pmb{x}}^i_{t-1}) = \pmb{d}^i_t$ as the output of each IterationModule that models an iteration $t$. 
From Eq.~\eqref{eq:ipfn-loss} we identify that the overall filtered trajectories inferred by IterativePFN will also be stochastic in nature as it uses an Adaptive Ground Truth (AGT), $\pmb{Y}_t=\pmb{Y}+\sigma_t\xi~\land~\xi\sim\mathcal{N}(0, I)$, within the training objective. Modelling the iterative filtering process internally results in a large network ($>$3.2M parameters). The AGT, while recovering the surface, causes clustering along the surface and fails to respect the original point distribution. This is illustrated in Fig.~\ref{fig:point-distribution}.

Inspired by Reflow which focuses on optimal transport between samples of different distributions~\cite{Liu-RectifiedFlow,Wu-PointStraightFlow}, we pose filtering as an optimal transport problem between point cloud patches. These patches of $n$ points are sampled from 1) the clean surface (i.e., clean patch $\pmb{X}_1\sim\pi_1$) and 2) a high noise variant of the clean surface (i.e., high noise patch $\pmb{X}_0\sim\pi_0$). The goal is to determine a transport plan (coupling) such that $\pmb{X}_1 = V(\pmb{X}_0)$ where $V:\mathbb{R}^{n\times 3}\rightarrow\mathbb{R}^{n\times 3}$. The approximated flow velocity, $\pmb{v}_\theta:\mathbb{R}^{n\times 3}\rightarrow\mathbb{R}^{n\times 3}$ satisfies this transport mapping and $(\Tilde{\pmb{X}}_0, \Tilde{\pmb{X}}_1)$ form a valid coupling. Consequently, the filtered trajectories are straight (see Fig.~\ref{fig:straight-traj-demo}), unlike ScoreDenoise and IterativePFN. Moreover, our method recovers the underlying point distribution of the surface without clustering artifacts or distortion. 
\section{Method}
\label{sec:method}
\begin{figure}[!tp]
    \centering
    \includegraphics[width=0.8\linewidth]{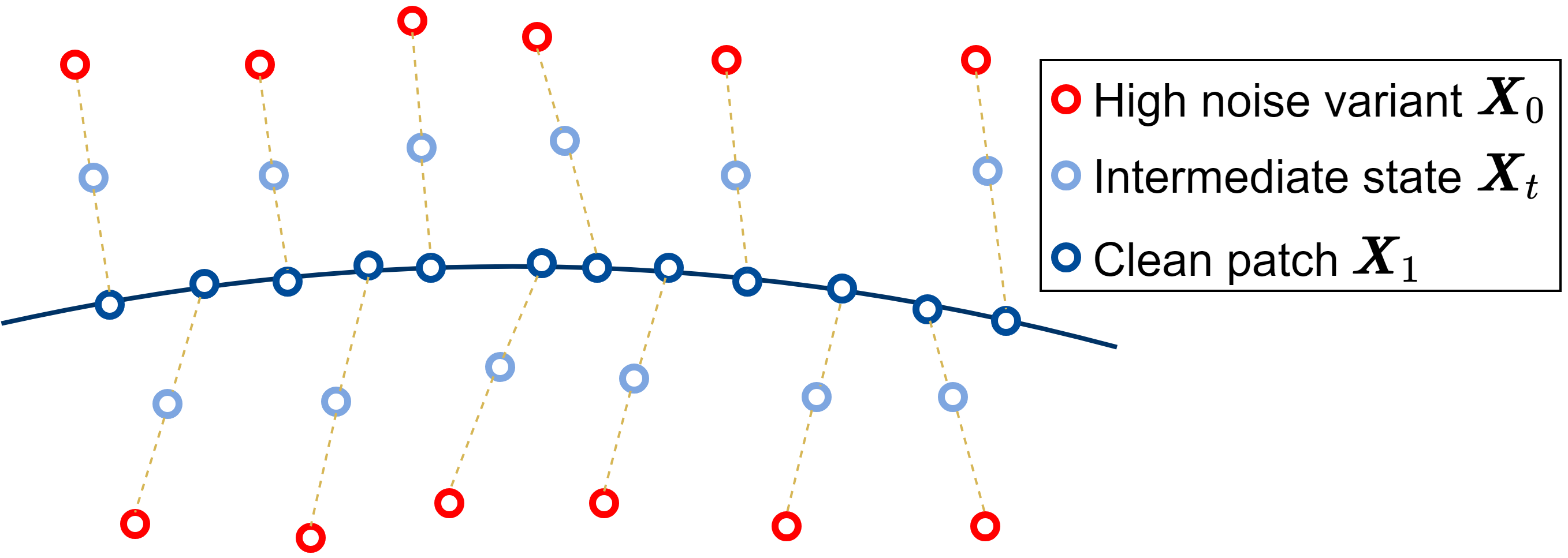}
    \caption{StraightPCF models initial noisy patches (light blue) as being intermediate states of a linear interpolation between high noise variants (red) and the clean surfaces (dark blue) and encourages straight filtering trajectories.}
    \label{fig:straight-traj-demo}
\end{figure}

Previous methods such as ScoreDenoise~\cite{Luo-Score-Based-Denoising} and IterativePFN~\cite{Edirimuni-IterativePFN} focus on filtering as a reverse Markov process where the forward process would correspond to adding noise. We formulate filtering as an optimal transport plan that moves noisy points back to the clean surface along the shortest (straight) path. We model noisy input patches as intermediate states between high noise variants and their corresponding clean counterparts. We design a graph-convolution based VelocityModule that infers a constant flow velocity for each intermediate state. This encourages noisy points to move along straight filtering trajectories as shown in Fig.~\ref{fig:straight-traj-demo}. We further improve the straightness of these trajectories via VelocityModule coupling. Finally, as the flow is constant, it may result in filtered points overshooting the clean surface. Therefore, we design a DistanceModule that scales the flow velocity appropriately and ensures convergence near the surface. The overall StraightPCF architecture is illustrated in Fig.~\ref{fig:straightpcf-network}, which demonstrates the filtering process that utilizes both the coupled VelocityModule and DistanceModule sub-networks to move points along straightened paths. The \textit{supplementary document} provides additional methodology details.

\subsection{Filtering via straight flows}
\label{subsec:straight-flows}
In this section, we introduce the VelocityModule (VM) that moves noisy points along constant, straight flows towards the clean surface. Given a noisy patch $\pmb{X}=\{\pmb{x}^i|\pmb{x}^i\in kNN(\pmb{x}^r, \mathcal{P}_{\pmb{X}}, k)\}$ centered around a reference point $\pmb{x}^r$ in the noisy point cloud $\mathcal{P}_{\pmb{X}}$, the filtering objective aims to move $\pmb{x}^i$ towards the underlying clean patch $\pmb{Y}=\{\pmb{y}^i|\pmb{y}^i\in \mathcal{P}_{\pmb{Y}}\}$. For the filtering task, learning based methods are typically trained on noisy point clouds with Gaussian noise~\cite{Rakotosaona-PCN,Zhang-Pointfilter,Luo-Score-Based-Denoising,Mao-PDFlow,Chen-DeepPSR}. Iterative filtering depends on the ability of the method to filter patches from a higher noise level to a lower noise level, with the end result converging to the clean surface. Given the highest noise setting $\sigma_H$, intermediate noise scales at a time $t\in[0,1]$ can be expressed as $\sigma = (1-t)\sigma_H$. Therefore, we model noisy patches $\pmb{X}=\pmb{Y}+\sigma \xi~\land~\xi\sim\mathcal{N}(0,I)$ as intermediate states $\pmb{X}_t$, of the filtering objective that moves points from a high noise variant, $\pmb{X}_0 = \pmb{Y} + \sigma_H \xi~\land~\xi\sim\mathcal{N}(0,I)$ to the clean counterpart $\pmb{X}_1 = \pmb{Y}$. Here, $\sigma_H=2\%$ of the point cloud's bounding sphere radius and corresponds to the highest noise setting of our training set. We observe that intermediate states $\pmb{X}_t$ can be defined as a linear interpolation of $\pmb{X}_0$ and $\pmb{X}_1$, that is, a straight path, such that,
\begin{equation}
\label{eq:linear-interpolation}
\pmb{X}_t = (1-t)\pmb{X}_0 + t\pmb{X}_1
\end{equation}

Therefore, we intuit that the filtering objective can be reformulated as an optimal transport process that moves points from $\pmb{X}_0$ to $\pmb{X}_1$. The flow of $\pmb{X}_t$ at time $t$ can be expressed via the ordinary differential equation (ODE):
\begin{equation}
\label{eq:ode}
\text{d}\pmb{X}_t = \pmb{v}(\pmb{X}_t)\text{d}t,
\end{equation}
where $\pmb{v}(\pmb{X}_t)$ is the velocity field at $\pmb{X}_t$. For this ODE process, the linear interpolation of Eq.~\eqref{eq:linear-interpolation} can be used in the least squares optimization to determine the flow,
\begin{equation}
\label{eq:ode-solution}
\min_{v}\int^1_0\mathbb{E}_{t\sim\mathcal{U}(0,1)}\left[\norm{\pmb{v}(\pmb{X}_t) - (\pmb{X}_1 - \pmb{X}_0)}^2_2\right]\text{d}t,
\end{equation}
where times $t$ are sampled uniformly along $[0,1]$. However, the velocity field at an intermediate state $\pmb{X}_t$ cannot be causally determined as both $\pmb{X}_0$ and $\pmb{X}_1$ are unknown, during filtering. To address this, our idea is to approximate the velocity field for each $\pmb{X}_t$ by a neural network.

\textbf{Training objective for single VM.} We use a DGCNN based graph neural network to model a VelocityModule $\pmb{v}_\theta$, with parameters $\theta$ to approximate the flow velocity. To train this VelocityModule, we define the following training objective.
\begin{equation}
\label{eq:vm-loss}
\mathcal{L}_A = \mathbb{E}_{t\sim\mathcal{U}(0,1)}\left[\norm{\pmb{v}_\theta(\pmb{X}_t) - \pmb{\delta}(\pmb{X}_1, \pmb{X}_0)}^2_2\right],
\end{equation}
where $\pmb{\delta}(\pmb{X}_1, \pmb{X}_0) = \pmb{X}_1 - \pmb{X}_0$. Hence, at each intermediate state, the corresponding flow velocity is ideally a constant vector that leads to a straight path from $\pmb{X}_0$ to $\pmb{X}_1$.

\textbf{Filtering objective for single VM.} During inference, given a straight flow velocity $\pmb{v}_\theta$, we can move points from $\pmb{X}_{\hat{t}/N}$ to $\pmb{X}_{(\hat{t}+1)/N}$ using the Euler method ODE solver: 
\begin{equation}
\label{eq:single-euler-step}
\Tilde{\pmb{X}}_{(\hat{t}+1)/N} = \Tilde{\pmb{X}}_{\hat{t}/N} + \frac{1}{N}\pmb{v}_\theta(\Tilde{\pmb{X}}_{\hat{t}/N}),
\end{equation}
where $N$ denotes the total number of filtering steps, $\hat{t}\in [0,\ldots, M,\ldots, N-1]$ is an integer time step and $\Tilde{\pmb{X}}_0=\pmb{X}_0$. In practice, filtering starts at an unknown time step $\hat{t}=M$. We do not know the noise scale of the input patch and need to model it as an intermediate state $\Tilde{\pmb{X}}_{M/N}$ of a higher noise variant $\pmb{X}_0$. The full position update takes the form,
\begin{equation}
\label{eq:full-euler-step}
\Tilde{\pmb{X}}_1 = \Tilde{\pmb{X}}_{M/N} + \frac{1}{N}\sum^{N-1}_{\hat{t}=M}\pmb{v}_\theta(\Tilde{\pmb{X}}_{\hat{t}/N}),
\end{equation}
Eq.~\eqref{eq:full-euler-step} poses two challenges:
\begin{enumerate}
    \item The straightness of paths is crucial to reducing the number of total steps and improving efficiency. If paths are not sufficiently straight (i.e., they are curved), we need a higher number of steps to filter points effectively.
    \item The starting time $t=M/N$ is unknown. Applying the Euler method for too many steps, $N$, may lead to points not converging at the surface.
\end{enumerate}

\midsepremove
\begin{table*}[!tp]
\setlength{\tabcolsep}{5pt}
\centering
\begin{tabular}{c|l| cccccc|cccccc}
\toprule
\multicolumn{2}{c|}{Resolution} & \multicolumn{6}{c|}{10K (Sparse)} & \multicolumn{6}{c}{50K (Dense)} \\
\midrule
\multicolumn{2}{c|}{Noise} & \multicolumn{2}{c}{1\%} & \multicolumn{2}{c}{2\%} & \multicolumn{2}{c|}{3\%} & \multicolumn{2}{c}{1\%} & \multicolumn{2}{c}{2\%} & \multicolumn{2}{c}{3\%} \\
\midrule
\multicolumn{2}{c|}{Method} & CD & P2M & CD & P2M & CD & P2M & CD & P2M & CD & P2M & CD & P2M \\
\midrule
\parbox[t]{2mm}{\multirow{7}{*}{\rotatebox[origin=c]{90}{PUNet dataset~\cite{Yu-PUNet}}}}
    & PCN~\cite{Rakotosaona-PCN} & 3.515 & 1.148 & 7.467 & 3.965 & 13.067 & 8.737 & 1.049 & 0.346 & 1.447 & 0.608 & 2.289 & 1.285  \\
    & PointFilter~\cite{Zhang-Pointfilter} & 2.461 & 0.443 & 3.534 & 0.862 & 5.089 & 1.849 & 0.758 & 0.182 & 0.907 & \uline{0.251} & 1.599 & 0.710 \\
    & Score~\cite{Luo-Score-Based-Denoising} & 2.521 & 0.463 & 3.686 & 1.074 & 4.708 & 1.942 & 0.716 & 0.150 & 1.288 & 0.566 & 1.928 & 1.041 \\
    & PDFlow~\cite{Mao-PDFlow} & 2.126 & 0.381 & 3.246 & 1.010 & 4.447 & 1.999 & 0.651 & 0.164 & 1.173 & 0.581 & 1.914 & 1.210 \\
    & DeepPSR~\cite{Chen-DeepPSR} & 2.353 & 0.306 & 3.350 & 0.734 & \uline{4.075} & \uline{1.242} & 0.649 &  \uline{0.076} & 0.997& 0.296 & \uline{1.344} & \bf 0.531 \\
    & IterativePFN~\cite{Edirimuni-IterativePFN} & \uline{2.056} & \bf 0.218 & \uline{3.043} & \bf 0.555 & 4.241 & 1.376 & \uline{0.605} & \bf 0.059 & \uline{0.803} & \bf 0.182 & 1.971 & 1.012  \\
\cmidrule{2-14}
    & \bf Ours & \bf 1.870 & \uline{0.239} & \bf 2.644 &\uline{ 0.604} & \bf 3.287 & \bf 1.126 & \bf 0.562 & 0.111 & \bf 0.765 & 0.266 & \bf 1.307 & \uline{0.648} \\  
\midrule
\parbox[t]{2mm}{\multirow{7}{*}{\rotatebox[origin=c]{90}{PCNet dataset~\cite{Rakotosaona-PCN}}}}
    & PCN~\cite{Rakotosaona-PCN} & 3.847 & 1.221 & 8.752 & 3.043 & 14.525 & 5.873 & 1.293 & 0.289 & 1.913 & 0.505 & 3.249 & 1.076  \\
    & Score~\cite{Luo-Score-Based-Denoising} &  3.369 &  0.830 &  5.132 &  1.195 &  6.776 &  1.941 &  1.066 &  0.177 &  1.659 &  0.354 &  2.494 &  0.657  \\
    & PointFilter~\cite{Zhang-Pointfilter} & 3.019 & 0.886 & 4.885 & 1.275 & 7.062 & 2.032 & 1.053 & 0.186 & 1.349 & \uline{0.257} & 2.225 &  \uline{0.491} \\
    & PDFlow~\cite{Mao-PDFlow} & 3.243 & \uline{0.606} & 4.545 & \uline{0.966} & \uline{5.934} & \uline{1.441} & 0.969 & 0.152 & 1.646 & 0.424 & 2.450 & 0.569 \\
    & DeepPSR~\cite{Chen-DeepPSR} &  2.873 & 0.783 & 4.757 & 1.118 & 6.031 & 1.619 & 1.010 & 0.146 & 1.515 & 0.340 & \uline{2.093} & 0.573 \\
    & IterativePFN~\cite{Edirimuni-IterativePFN} & \bf 2.621 & 0.698 & \uline{4.439} & 1.011 & 6.026 & 1.560 & \uline{0.913} & \bf 0.139 & \uline{1.251} & \bf 0.238 & 2.529 & 0.716 \\
\cmidrule{2-14}
    & \bf Ours & \uline{2.747} & \bf 0.536 & \bf 4.046 & \bf 0.788 & \bf 4.921 & \bf 1.093 & \bf 0.877 & \bf 0.144 & \bf 1.173 & \uline{0.259} & \bf 1.816 & \bf 0.445 \\
\bottomrule
\end{tabular}
\caption{Quantitative filtering results of recent state-of-the-art methods and our method on the synthetic PUNet and PCNet datasets. Note that our network is lightweight, with just $\sim 530K$ parameters (17\% of IterativePFN). CD and P2M values are multiplied by $10^4$. }
\label{tab:pu-pc-results}
\end{table*}
\midsepdefault

\subsection{Straighter flows via VelocityModule coupling}
\label{subsec:vm-coupling}
Given noisy initial data, the trajectories of points tend to be curved, with limited straightness. One way to improve straightness is to finetune the velocity network on the coupling $(\Tilde{\pmb{X}}_0, \Tilde{\pmb{X}}_1)$ to satisfy $\Tilde{\pmb{X}}_1 = V(\Tilde{\pmb{X}}_0)$. For filtering, we aim to recover surfaces while preserving local geometric details~\cite{Rakotosaona-PCN}. Applying such 
finetuning requires the pre-computation of $\Tilde{\pmb{X}}_1$ for all possible surface patches and is infeasible due to the large number of patches in the training set.
We propose a simple mechanism to straighten paths by coupling $K$ VelocityModules together.

\textbf{Training objective for coupled VMs.} Given a noisy patch $\pmb{X}_t$, we partition the trajectory from $\pmb{X}_t$ to $\pmb{X}_1$ into $K$ segments and obtain the velocity flow $\pmb{v}^k_\theta(\Tilde{\pmb{X}}_{t_k})$ at times $t_k=(t(K-k)+k)/K$ where $k\in\{0,1,\ldots,K-1\}$. Intermediate positions at times $t_{k+1}$ are given by $\Tilde{\pmb{X}}_{t_{k+1}} = \Tilde{\pmb{X}}_{t_k} + (1/K)~\pmb{v}^k_\theta(\Tilde{\pmb{X}}_{t_k})$.
We empirically find that two VelocityModules, i.e., $K=2$, provide the best balance between accuracy and efficiency. The VelocityModules are pretrained using the training objective of Eq.~\eqref{eq:vm-loss} and the coupled VMs are finetuned with the introduced objective,
\begin{equation}
\label{eq:coupled_vm-loss}
\begin{split}
\mathcal{L}_B = \mathbb{E}_{t\sim\mathcal{U}(0,1)}\left[\sum^{K-1}_{k=0}\norm{\pmb{v}^k_\theta(\Tilde{\pmb{X}}_{t_k}) - \pmb{\delta}(\pmb{X}_1, \pmb{X}_0)}^2_2 \right. \\
\left. + \lambda_1\sum^{K-2}_{k=0}\norm{\pmb{\delta}(\Tilde{\pmb{X}}_{t_{k+1}}, \pmb{X}_{t_{k+1}})}^2_2\right],
\end{split}
\end{equation}
for $K\geq2$, $\lambda_1=10$ and $\Tilde{\pmb{X}}_{t_0}=\pmb{X}_{t_0}=\pmb{X}_t$. The first term of $\mathcal{L}_B$ encourages coupled VelocityModules to infer a constant velocity at times $t_k$, while the second encourages filtered points $\Tilde{\pmb{X}}_{t_k}$ to move closer to interpolated points $\pmb{X}_{t_k}$.

\textbf{Filtering objective for coupled VMs.} At inference, we apply a modified form of the Euler method to filter points, similar to Eq.~\eqref{eq:single-euler-step}. We apply $K$ coupled VelocityModules $N$ times, resulting in $T=K\cdot N$ total filtering steps. We adjust our earlier notation of integer time steps such that $\hat{t}\in\{0, K,\ldots \hat{M},\ldots,K(N-1)\}$ where $\hat{M}$ is divisible by $K$ and $\Tilde{\pmb{X}}_0 = \pmb{X}_0$. The sequential position update becomes,
\begin{equation}
\label{eq:modified-straight-single-euler-step}
\Tilde{\pmb{X}}_{(\hat{t}+k+1)/T} = \Tilde{\pmb{X}}_{(\hat{t}+k)/T} + \frac{1}{T}\pmb{v}^k_\theta(\Tilde{\pmb{X}}_{(\hat{t}+k)/T}).
\end{equation}  
The corresponding full position update across time steps $\mathcal{S}=\{\hat{M}, \hat{M} + K, \ldots, N(K-1)\}$ is,
\begin{equation}
\label{eq:modified-full-euler-step}
\Tilde{\pmb{X}}_1 = \Tilde{\pmb{X}}_{\hat{M}/T} + \frac{1}{T}\sum_{\hat{t}\in \mathcal{S}}\sum^{K-1}_{k=0}\pmb{v}^k_\theta(\Tilde{\pmb{X}}_{(\hat{t}+k)/T}).
\end{equation}  

\subsection{Distance estimation to the clean surface}
\label{subsec:distance-estimation}
\begin{figure}[!tp]
    \centering
    \includegraphics[width=0.98\linewidth]{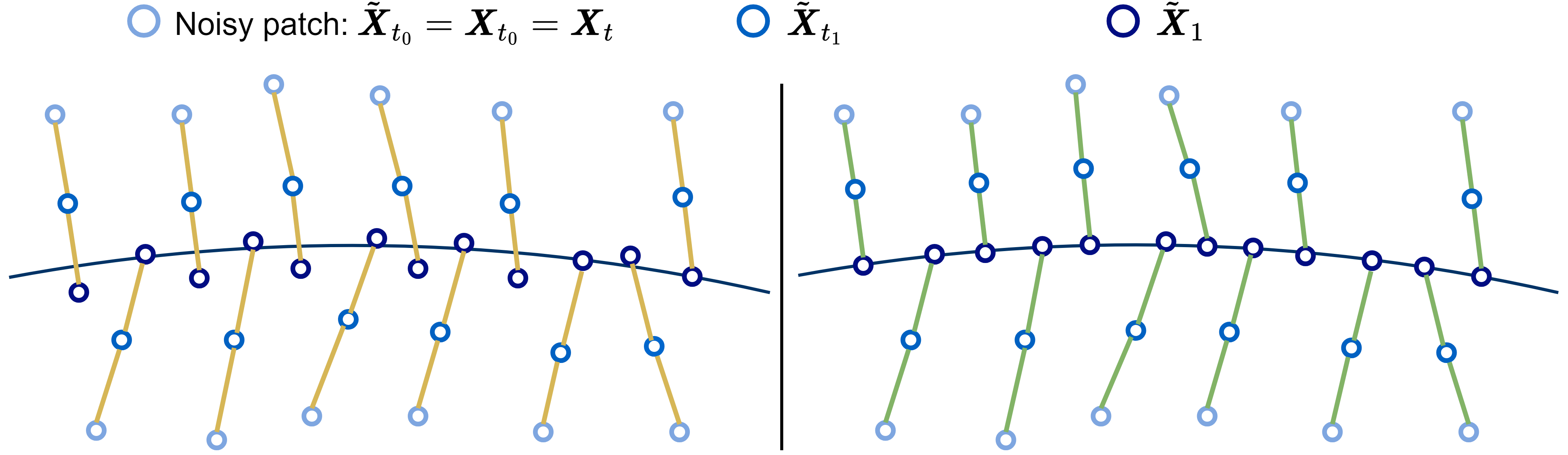}
    \caption{Left: Filtering by coupled VelocityModules only. Right: Coupled VelocityModules and DistanceModule. Scaled trajectories (green lines) lead to better convergence at the surface.}
    \label{fig:straight-traj-demo-with-dm}
\end{figure}

We now address the second challenge related to Eq.~\eqref{eq:full-euler-step} and introduce the DistanceModule that scales the overall straight trajectory. This leads to better convergence near the surface. The DistanceModule estimates a  distance scalar, corresponding to the standard deviation of initial noisy points from the clean surface. More specifically, the DistanceModule $\mathcal{D}_\phi(\cdot)$ is used to approximate a mapping $d_\phi:\mathbb{R}^{n\times 3}\rightarrow \mathbb{R}$, such that,
\begin{equation}
\label{eq:distance-module-output}
d_\phi(\Tilde{\pmb{X}}_{t_0}) = \text{Sigmoid}(\text{Max}(\mathcal{D}_\phi(\Tilde{\pmb{X}}_{t_0}))). 
\end{equation}
The distance $d_\phi$ is then used to scale the output of VelocityModules, as illustrated in Fig.~\ref{fig:straight-traj-demo-with-dm}. The DistanceModule is embedded within the StraightPCF architecture (see Fig.~\ref{fig:straightpcf-network}). We keep the weights of the finetuned coupled VelocityModules fixed when training the DistanceModule. The respective training objective for optimizing the parameters $\phi$ is,
\begin{equation}
    \begin{split}
        \label{eq:dm-loss}
        \mathcal{L}_{C} = \mathbb{E}_{t\sim\mathcal{U}(0,1)}\left[\left(d_\phi(\Tilde{\pmb{X}}_{t_0}) - \frac{\norm{\pmb{\delta}(\pmb{X}_1, \pmb{X}_{t_0})}_2}{\norm{\pmb{\delta}(\pmb{X}_1, \pmb{X}_0)}_2}\right)^2 \right. \\ 
        \left. \vphantom{\left(d_\phi(\pmb{X}_{t_0}) - \frac{\norm{\pmb{\delta}(\pmb{X}_1, \pmb{X}_{t_0})}_2}{\norm{\pmb{\delta}(\pmb{X}_1, \pmb{X}_0)}_2}\right)^2} + \lambda_2\norm{\pmb{\delta}(\Tilde{\pmb{X}}_1, \pmb{X}_1)}^2_2\right],
    \end{split}
\end{equation}
where $t_0=t$ and $\Tilde{\pmb{X}}_{t_0}=\pmb{X}_{t_0}=\pmb{X}_t$. The first term of $\mathcal{L}_C$ encourages the DistanceModule to infer the relative distance of $\pmb{X}_{t_0}$ from the clean surface, as compared to $\pmb{X}_0$. The last term encourages points to return to the clean surface. We set $\lambda_2=2\times10^2$ to ensure the loss contribution of the last term is of the same order as that of the first one. Consequently, Eq.~\eqref{eq:modified-straight-single-euler-step} becomes,
\begin{equation}
\label{eq:eq:modified-straight-single-euler-step-with-dm}
\Tilde{\pmb{X}}_{(\hat{t}+k+1)/T} = \Tilde{\pmb{X}}_{(\hat{t}+k)/T} + \frac{d_\phi(\Tilde{\pmb{X}}_{\hat{M}/T})}{T}\pmb{v}^k_\theta(\Tilde{\pmb{X}}_{(\hat{t}+k)/T}).
\end{equation}

\begin{figure*}[!tp]
    \centering
    \includegraphics[width=\linewidth]{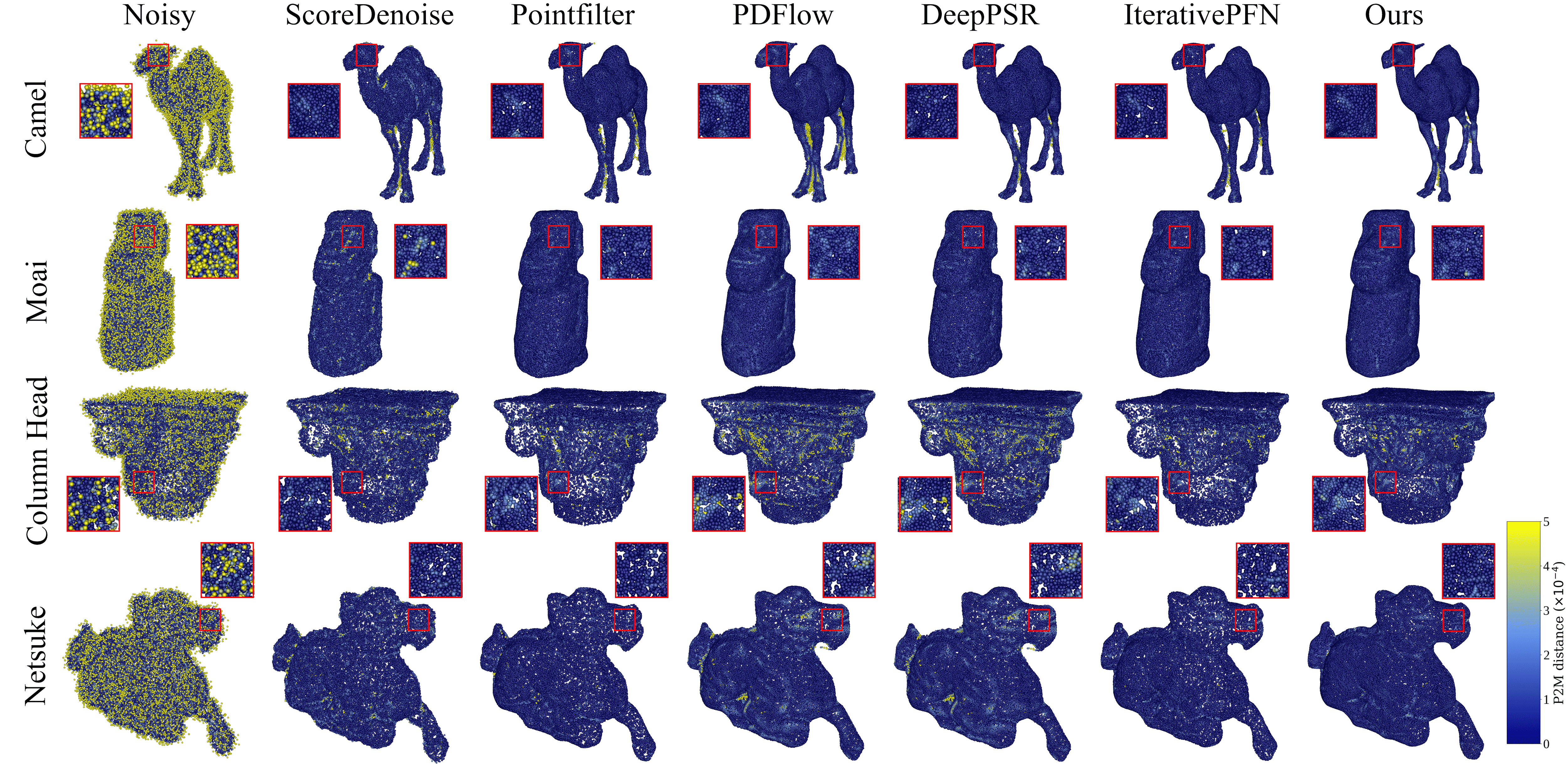}
    \caption{Visual filtering results for 50K resolution shapes ($\sigma=2\%$) within the PUNet and PCNet datasets. The darker (i.e., more blueish) the better. We achieve both strong point-wise P2M results while also ensuring well distributed points (illustrated by close-ups) unlike DeepPSR or IterativePFN which have holes indicating clustering.}
    \label{fig:pu-pc-results}
\end{figure*}

\section{Experiments}
\label{sec:results}
Next, we provide results on synthetic data under Gaussian noise and real-world Kinect~\cite{Wang-Kinect} and Paris-Rue-Madame~\cite{Serna-Rue-Madame} data. The \textit{supplementary document} contains additional results of both conventional and learning based methods on synthetic and real-world scanned data, as well as a comparison of testing times.

\subsection{Training and evaluation details}
\label{subsec:dataset}
We follow the training procedure of ScoreDenoise~\cite{Luo-Score-Based-Denoising}, and train our model on the PUNet dataset~\cite{Yu-PUNet} consisting of 40 point clouds for training and 20 point clouds for testing. To ensure consistency with ScoreDenoise's training settings, we add Gaussian noise sampled with standard deviation $\sigma_H=2\%$ of the bounding sphere radius to each clean point cloud. Our training procedure only requires the high noise variants and the clean ground truth targets. All intermediate states at noise scales $\sigma=(1-t)\sigma_H$, with $t\sim\mathcal{U}(0,1)$, are created as linear interpolations between these two, initial and final states, as per Eq.~\eqref{eq:linear-interpolation}. For testing, we also consider 10 test point clouds from the PCNet dataset~\cite{Rakotosaona-PCN} provided by ScoreDenoise. These synthetic point clouds contain Gaussian noise at scales of $1\%$, $2\%$ and $3\%$ of the point cloud's bounding sphere radius. We use two different sampling densities of 10K and 50K points to evaluate filtering ability across different sparsity settings. Moreover, to assess filtering results on real-world noisy point clouds, we compare methods on the Kinect v1 dataset that comprises of 71 point clouds~\cite{Wang-Kinect} and 4 scenes extracted from the Paris-Rue-Madame dataset~\cite{Serna-Rue-Madame}. Finally, we provide a comparison on 4 scenes of the Kitti-360 dataset~\cite{Liao-Kitti360}, in the \textit{supplementary document}. Following~\cite{Luo-Score-Based-Denoising, Edirimuni-IterativePFN}, all methods are only trained on PUNet with Gaussian noise.

\textbf{Implementation}. We train and test StraightPCF on a NVIDIA GeForce RTX 3090 GPU using PyTorch. We use the Adam optimizer with a learning rate of $1\times 10^{-4}$. Similar to~\cite{Luo-Score-Based-Denoising, Edirimuni-IterativePFN}, we use PyTorch3D~\cite{Ravi-PyTorch3D} to compute Chamfer Distance (CD) and Point2Mesh (P2M) metric values.

\subsection{Performance on synthetic data} 
Table~\ref{tab:pu-pc-results} and Fig.~\ref{fig:pu-pc-results} illustrate filtering results on the synthetic PUNet and PCNet datasets. Our method has a clear edge over others in recovering the underlying clean point distribution demonstrated by superior CD results across all noise settings. We also achieve strong P2M results on the PUNet dataset. While IterativePFN~\cite{Edirimuni-IterativePFN} has marginally better P2M results on the PUNet data, this does not ensure a good distribution of points as evidenced by visual results in Fig.~\ref{fig:pu-pc-results}. Here, for noisy point clouds at 50K resolution and 2\% noise, IterativePFN exhibits small holes, indicating clustering. Moreover, our network consists of only 530K parameters and is roughly 83\% smaller than that of IterativePFN, given the latter's 3.2M parameters. DeepPSR~\cite{Chen-DeepPSR}, despite using a post-processing regularization step, also suffers from holes and poor point distributions. We also note that IterativePFN generalizes poorly to higher noise scales outside the training noise scales (where maximum noise is $\sigma=2\%$). The ability of our method to generalize well across different data is demonstrated by the superior performance on the PCNet dataset. The PCNet shapes are unseen during training. Our method obtains both superior CD and P2M results over all other methods. At 10K resolution and $\sigma=3\%$ noise, our method offers a 17.1\% reduction in CD error and 24.1\% reduction in P2M error. Our method also yields state-of-the-art results at 50K resolution where, for $\sigma=3\%$, the CD error improvement is 13.2\% and P2M error improvement is 9.4\%. Visual filtering results in Fig.~\ref{fig:pu-pc-results} indicate that our method outperforms others in recovering complex details such as the legs of the Camel which are closely situated. By contrast, methods such as Pointfilter~\cite{Zhang-Pointfilter} and PDFlow~\cite{Mao-PDFlow} flatten points between these close surfaces while IterativePFN does not recover the body well. Furthermore, on a complex shape as Netsuke, we outperform other methods which either cause clustering (e.g., PDFlow, IterativePFN) at this high noise scale or leave behind noise due to limited filtering ability (e.g., ScoreDenoise).

\begin{figure*}[!tp]
    \centering
    \includegraphics[width=0.98\linewidth]{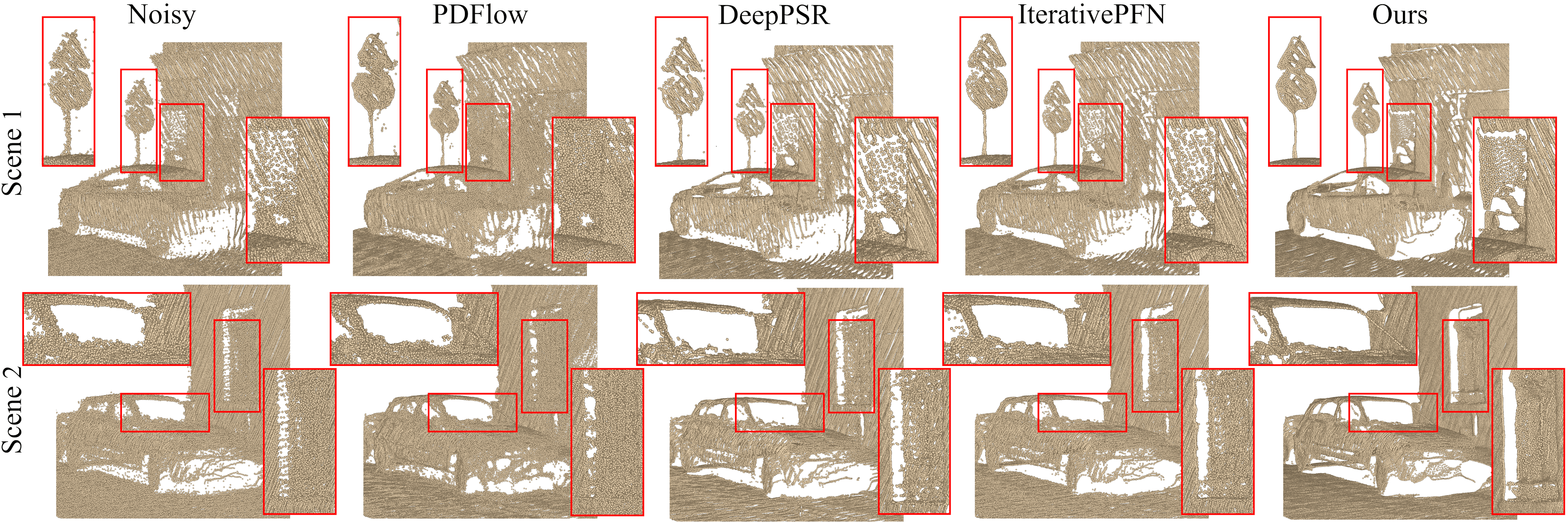}
    \caption{Visual filtering results on two scenes of the RueMadame dataset. We show results for the most recent state-of-the-art methods.}
    \label{fig:ruemadame-results}
\end{figure*}

\subsection{Performance on real-world scanned data}
Next we consider filtering results on scanned data. Table~\ref{tab:kinect-results} provides quantitative results on the Kinect dataset and Fig.~\ref{fig:ruemadame-results} illustrates visual results on the Paris-Rue-Madame dataset. Our method performs favorably on the Kinect data, obtaining a 1.82\% reduction on CD errors, as compared to IterativePFN~\cite{Edirimuni-IterativePFN}. As this data is very sparse, our P2M result is marginally higher than Pointfilter~\cite{Zhang-Pointfilter} and ScoreDenoise~\cite{Luo-Score-Based-Denoising}. These methods focus on returning points to the surface yet succumb to clustering artifacts whereas our method both recovers surfaces while retaining good point distributions. The Paris-Rue-Madame dataset contains high noise scans due to outdoor environmental factors. In Fig.~\ref{fig:ruemadame-results}, we visualize filtering results for several most recent, state-of-the-art methods. In general, methods such as PDFlow~\cite{Mao-PDFlow} and DeepPSR~\cite{Chen-DeepPSR} perform poorly in removing noisy artifacts while IterativePFN~\cite{Edirimuni-IterativePFN} causes points to cluster near the original scan lines. Our method, however, recovers the underlying clean surface and ensures a much better point distribution, as evidenced by the close-ups in Scene 1. In Scene 2, PDFlow and IterativePFN are not able to clean the surface of the parked vehicle whereas our method recovers the underlying shape with fewer noisy artifacts.

\midsepremove
\begin{table}[!tp]
\centering
\begin{tabular}{l|ll}
\toprule
\multicolumn{1}{c|}{\multirow{2}{*}{Method}} & \multicolumn{2}{c}{Kinect} \\
\cmidrule{2-3}
 & CD & P2M \\
\midrule
ScoreDenoise & 1.322 & {\ul 0.652} \\
Pointfilter & 1.377 & \textbf{0.644} \\
PDFlow & 1.334 & 0.699 \\
DeepPSR & 1.431 & 0.715 \\
IterativePFN & {\ul 1.320} & 0.685 \\
\midrule
Ours & \textbf{1.296} & 0.664 \\
\bottomrule
\end{tabular}
\caption{Quantitative results on Kinect data. Our network is lightweight, with just $\sim 530K$ parameters (17\% of IterativePFN). CD and P2M values are multiplied by $10^4$.}
\label{tab:kinect-results} 
\end{table}
\midsepdefault

\section{Ablation Study and Discussion}
\label{sec:ablation}
We train and test several variant architectures in order to ascertain the impact of the proposed VelocityModule (VM) and DistanceModule (DM). The results are given in Table~\ref{tab:ablation}. Our coupled VM + DM architecture (V5) increases the parameter number to $\sim 530K$. Therefore, we train another Large VM with $\sim 530K$ parameters (V3). We find that directly increasing the parameter number (V3) leads to very limited performance gain while V5 exhibits far superior performance. Furthermore, the coupled VM + DM architecture (V5) significantly outperforms the single VM (V1 and V2) architectures. Finally, there is a noticeable difference in performance with and without a DM as evidenced by the disparity between V4 and V5. We provide further ablations, including higher VM couplings, in the \textit{supplementary document}. 

\midsepremove
\begin{table}[!tp]
\centering
\setlength\tabcolsep{4pt} 
\begin{tabular}{l|ll|ll|ll}
\toprule
\multicolumn{1}{c|}{\multirow{3}{*}{Ablation}} & \multicolumn{6}{c}{10K points} \\
\cmidrule{2-7}
& \multicolumn{2}{c|}{1\% noise} & \multicolumn{2}{c|}{2\% noise} & \multicolumn{2}{c}{3\% noise} \\
\cmidrule{2-7}
& CD & P2M & CD & P2M & CD & P2M \\ 
\midrule
\textit{V1}) VM w/o DM & 2.16 & 0.42 & 3.06 & 0.92 & 3.73 & 1.42 \\
\textit{V2}) VM w/ DM & 2.00 & 0.32 & 3.06 & 0.96 & 3.81 & 1.55 \\ 
\textit{V3}) Large VM & 2.17 & 0.41 & 2.99 & 0.84 & 3.54 & 1.29 \\ 
\textit{V4}) CVM w/o DM & 1.97 & 0.32 & 3.01 & 0.92 & 3.71 & 1.45 \\ 
\midrule
\textbf{\textit{V5})} \textbf{CVM w/ DM} & \bf 1.87 & \bf 0.24 & \bf 2.64 &\bf 0.60 & \bf 3.29 & \bf 1.13  \\
\bottomrule
\end{tabular}
\caption{Ablation results for different VelocityModule (VM) and DistanceModule (DM) configurations.}
\label{tab:ablation} 
\end{table}
\midsepdefault

\textbf{Limitation.} While our StraightPCF yields state-of-the-art results across multiple datasets, we observe relatively low performance on low density or high sparsity data, which is similar to current methods. We provide the visual results in the \textit{supplementary document} due to space limit.

\section{Conclusion}
\label{sec:conslusion}
Recent deep learning based filtering methods focus on moving noisy points along stochastic paths to remove noise from input point clouds. We propose the first study to consider filtering points along straight paths, leading to smaller discretization errors and fewer filtering iterations. This lightweight method, while being parameter efficient, delivers filtered point distributions closer to that of the ground truth distributions without requiring any regularization in loss function or post-processing. Our method achieves state-of-the-art performance on multiple synthetic and real-world datasets across standard filtering metrics, showcasing its superiority and effectiveness. 

{
    \small
    \bibliographystyle{ieeenat_fullname}
    \bibliography{main}
}

\input{supplementary}

\end{document}

%% file: preamble.tex
\usepackage[dvipsnames]{xcolor}


%% file: supplementary.tex
\clearpage
\renewcommand\thesection{\Alph{section}}
\renewcommand\thesubsection{\thesection.\arabic{subsection}}
\setcounter{page}{1}
\setcounter{section}{0}
\maketitlesupplementary

This supplementary document contains the following:
\begin{enumerate}[label*=\Alph*.]
\item Additional methodology details.
    \begin{enumerate}[label*=\arabic*.]
        \item VelocityModule training.
        \item Inference time filtering objective for full network.
    \end{enumerate}
\item Further evaluation on synthetic and scanned data 
    \begin{enumerate}[label*=\arabic*.]
        \item Performance of additional methods on PUNet and PCNet data with Gaussian noise.
        \item Additional visual results on real-world scanned data.
    \end{enumerate}
\item Further evaluation on PUNet and PCNet data under different noise patterns.
\item Comparison of test times for different methods.
\item Further ablation studies.
    \begin{enumerate}[label*=\arabic*.]
        \item Higher VelocityModule couplings, $K$.
        \item Impact of Euler step number, $N$.
    \end{enumerate}
\item Discussion of limitations.
\end{enumerate}

\section{Additional Methodology Details}
\label{sec:additional-methodology}

\subsection{VelocityModule training}
\label{subsec:vm-training}
In Sec.~4.1 of the main paper, we presented the VelocityModule that enables filtering via straight flows. Here, we provide more detail into its training objective. We train the VelocityModule by minimizing the loss of Eq.~(7) in the main paper. During training, we first draw samples $\pmb{X}_0=\pmb{Y}+\sigma_H\xi~\land~\xi\sim\mathcal{N}(0,I)$ and $\pmb{X}_1=\pmb{Y}$ from $\pi_0$ and $\pi_1$, respectively. Given a pair $(\pmb{X}_0$, $\pmb{X}_1)$, we use the linear interpolation of Eq.~(4), given in the main paper, to sample an intermediate state at time $t\sim\mathcal{U}(0,1)$. For the filtering task, we do not have explicit knowledge of the time step at which we start the filtering process. Consequently, we do not input the time step $t$ to the VelocityModule, we only provide $\pmb{X}_t$. We then obtain the optimal parameters $\theta^\star$ of the VelocityModule by minimizing the expected value over time of the $L_2$ norm term in Eq.~(7), given in the main paper, such that,
\begin{equation}
\label{eq:vm-loss-full-form}
\theta^\star = \argmin_{\theta}\left\{\mathbb{E}_{t\sim\mathcal{U}(0,1)}\left[\norm{\pmb{v}_\theta(\pmb{X}_t) - \pmb{\delta}(\pmb{X}_1, \pmb{X}_0)}^2_2\right]\right\}.
\end{equation}

\subsection{Inference time filtering objective for full network}
\label{subsec:inference-details}
In this section, we provide supplementary details on the filtering objective of the full network, with trained VelocityModule coupling and DistanceModule. Eq. (15) of the main paper provides the sequential position update for our StraightPCF network. Now, the full position update across time steps $\mathcal{S}=\{\hat{M}, \hat{M} + K, \ldots, N(K-1)\}$ is,
\begin{equation}
\label{eq:final-full-euler-step}
\Tilde{\pmb{X}}_1 = \Tilde{\pmb{X}}_{\hat{M}/T} + \frac{d_\phi(\Tilde{\pmb{X}}_{\hat{M}/T})}{T}\sum_{\hat{t}\in \mathcal{S}}\sum^{K-1}_{k=0}\pmb{v}^k_\theta(\Tilde{\pmb{X}}_{(\hat{t}+k)/T}).
\end{equation}  
We apply our StraightPCF network across $T=N\cdot K$ total iterations. At very high noise levels, this entire filtering process has to be repeated, similar to ScoreDenoise~\cite{Luo-Score-Based-Denoising}. The reason for this is the upper limit of our training noise scales is only $\sigma=2\%$. Therefore, at inference, for $\sigma=2\%$ noise, we repeat the full position update $2$ times while for $\sigma=3\%$, we repeat the process $3$ times.

\midsepremove
\begin{table*}[!tp]
\setlength{\tabcolsep}{5pt}
\centering
\begin{tabular}{c|l| cccccc|cccccc}
\toprule
\multicolumn{2}{c|}{Resolution} & \multicolumn{6}{c|}{10K (Sparse)} & \multicolumn{6}{c}{50K (Dense)} \\
\midrule
\multicolumn{2}{c|}{Noise} & \multicolumn{2}{c}{1\%} & \multicolumn{2}{c}{2\%} & \multicolumn{2}{c|}{3\%} & \multicolumn{2}{c}{1\%} & \multicolumn{2}{c}{2\%} & \multicolumn{2}{c}{3\%} \\
\midrule
\multicolumn{2}{c|}{Method} & CD & P2M & CD & P2M & CD & P2M & CD & P2M & CD & P2M & CD & P2M \\
\midrule
\parbox[t]{2mm}{\multirow{7}{*}{\rotatebox[origin=c]{90}{PUNet dataset~\cite{Yu-PUNet}}}}
    & Bilateral~\cite{Fleishman-Bilateral} & 3.646 & 1.342 & 5.007 & 2.018 & 6.998 & 3.557 & 0.877 & 0.234 & 2.376 & 1.389 & 6.304 & 4.730 \\
    & Jet~\cite{Cazals-Jet} & 2.712 & 0.613 & 4.155 & 1.347 & 6.262 & 2.921 & 0.851 & 0.207 & 2.432 & 1.403 & 5.788 & 4.267 \\
    & MRPCA~\cite{Mattei-MRPCA} & 2.972 & 0.922 & 3.728 & 1.117 & 5.009 & 1.963 &  0.669 &  0.099 & 2.008 & 1.033 & 5.775 & 4.081 \\
    & GLR~\cite{Hu-GLR} & 2.959 & 1.052 & 3.773 & 1.306 & 4.909 & 2.114 & 0.696 & 0.161 & 1.587 & 0.830 & 3.839 & 2.707  \\
\cmidrule{2-14}
    & GPDNet\cite{Pistilli-GPDNet} & 3.780 & 1.337 & 8.007 & 4.426 & 13.482 & 9.114 & 1.913 & 1.037 & 5.021 & 3.736 & 9.705 & 7.998  \\
    & DMRD~\cite{Luo-DMRDenoise} & 4.482 & 1.722 & 4.982 & 2.115 & 5.892 & 2.846 & 1.162 & 0.469 & 1.566 & 0.800 & 2.432 & 1.528  \\
\cmidrule{2-14}
    & \bf Ours & \bf 1.870 & \bf 0.239 & \bf 2.644 & \bf 0.604 & \bf 3.287 & \bf 1.126 & \bf 0.562 & \bf 0.111 & \bf 0.765 & \bf 0.266 & \bf 1.307 & \bf 0.648 \\  
\midrule
\parbox[t]{2mm}{\multirow{7}{*}{\rotatebox[origin=c]{90}{PCNet dataset~\cite{Rakotosaona-PCN}}}}
    & Bilateral~\cite{Fleishman-Bilateral} & 4.320 & 1.351 & 6.171 & 1.646 & 8.295 & 2.392 & 1.172 & 0.198 & 2.478 & 0.634 & 6.077 & 2.189  \\
    & Jet~\cite{Cazals-Jet}  & 3.032 &  0.830 & 5.298 & 1.372 & 7.650 & 2.227 & 1.091 & 0.180 & 2.582 & 0.700 & 5.787 & 2.144  \\
    & MRPCA~\cite{Mattei-MRPCA} & 3.323 & 0.931 &  4.874 & 1.178 &  6.502 & 1.676 & 0.966 & 0.140 & 2.153 & 0.478 & 5.570 & 1.976 \\
    & GLR~\cite{Hu-GLR} & 3.399 & 0.956 & 5.274 &  1.146 & 7.249 &  1.674 & 0.964 & \bf 0.134 & 2.015 & 0.417 & 4.488 & 1.306  \\
\cmidrule{2-14}
    & GPDNet~\cite{Pistilli-GPDNet} & 5.470 & 1.973 & 10.006 & 3.650 & 15.521 & 6.353 & 5.310 & 1.716 & 7.709 & 2.859 & 11.941 & 5.130 \\
    & DMRD~\cite{Luo-DMRDenoise} & 6.602 & 2.152 & 7.145 & 2.237 & 8.087 & 2.487 & 1.566 & 0.350 & 2.009 & 0.485 & 2.993 & 0.859  \\
\cmidrule{2-14}
    & \bf Ours & \bf 2.747 & \bf 0.536 & \bf 4.046 & \bf 0.788 & \bf 4.921 & \bf 1.093 & \bf 0.877 & \uline{0.144} & \bf 1.173 & \bf 0.259 & \bf 1.816 & \bf 0.445 \\
\bottomrule
\end{tabular}
\caption{Quantitative filtering results of conventional methods and older learning based methods on the synthetic PUNet and PCNet datasets with Gaussian noise. CD and P2M values are multiplied by $10^4$. }
\label{tab:supp-pu-pc-results}
\end{table*}
\midsepdefault

\begin{figure*}[!tp]
\centering
\includegraphics[width=0.97\linewidth]{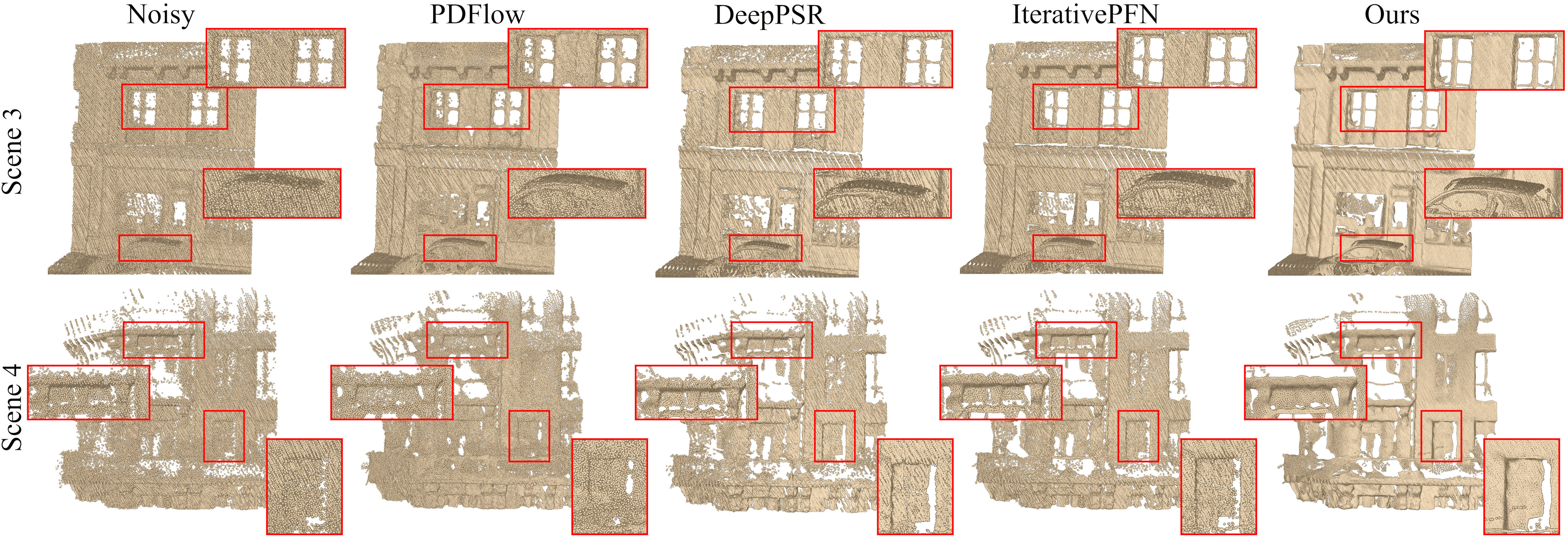}
\caption{Additional visual results on the real-world Paris-Rue-Madame dataset.}
\label{fig:supp-rm-results}
\end{figure*}

\section{Further Evaluation on Synthetic and Scanned Data }
\label{sec:further-results}
In this section, we provide additional results on both synthetic and scanned data that could not be added to the main paper, due to constraints of space. Moreover, we provide more detail into our experimental set-up. To obtain Chamfer Distance (CD) and Point2Mesh (P2M) results, we have used the implementation of PyTorch3D~\cite{Ravi-PyTorch3D} with the same settings as ScoreDenoise~\cite{Luo-Score-Based-Denoising}, PDFlow~\cite{Mao-PDFlow} and DeepPSR~\cite{Chen-DeepPSR}. For all filtering results, to ensure fair comparison, learning based methods were only trained on the synthetic PUNet training set with Gaussian noise.

\subsection{Performance of additional methods on PUNet and PCNet data with Gaussian noise}
Table~\ref{tab:supp-pu-pc-results} presents the performance of deep learning-based methods and conventional methods which were not included in the main paper, due to limitations of space. These results are for PUNet and PCNet data with Gaussian noise at scales of $1\%$, $2\%$ and $3\%$ of the bounding sphere's radius. In general, we see that these methods perform sub-optimally, with relatively higher CD and P2M errors. Furthermore, conventional methods require hyper-parameter tuning to obtain the best possible results. This is a tedious process which deep learning based methods help remedy.

\begin{figure*}[!tp]
\centering
\includegraphics[width=0.97\linewidth]{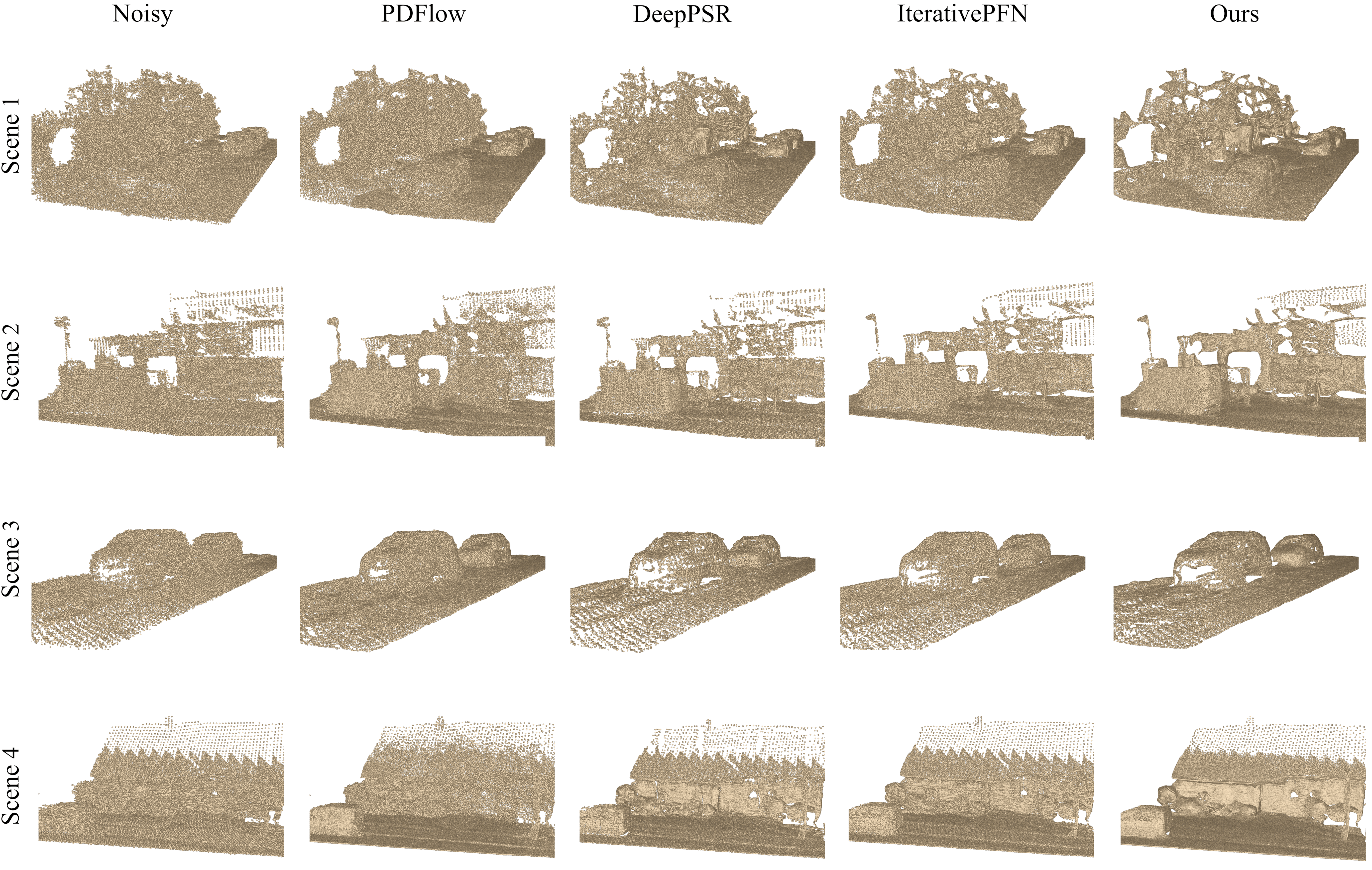}
\caption{Visual results on 4 scans of the real-world Kitti-360 data.}
\label{fig:supp-kitti-results}
\end{figure*}

\begin{figure*}[!tp]
\centering
\includegraphics[width=0.8\linewidth]{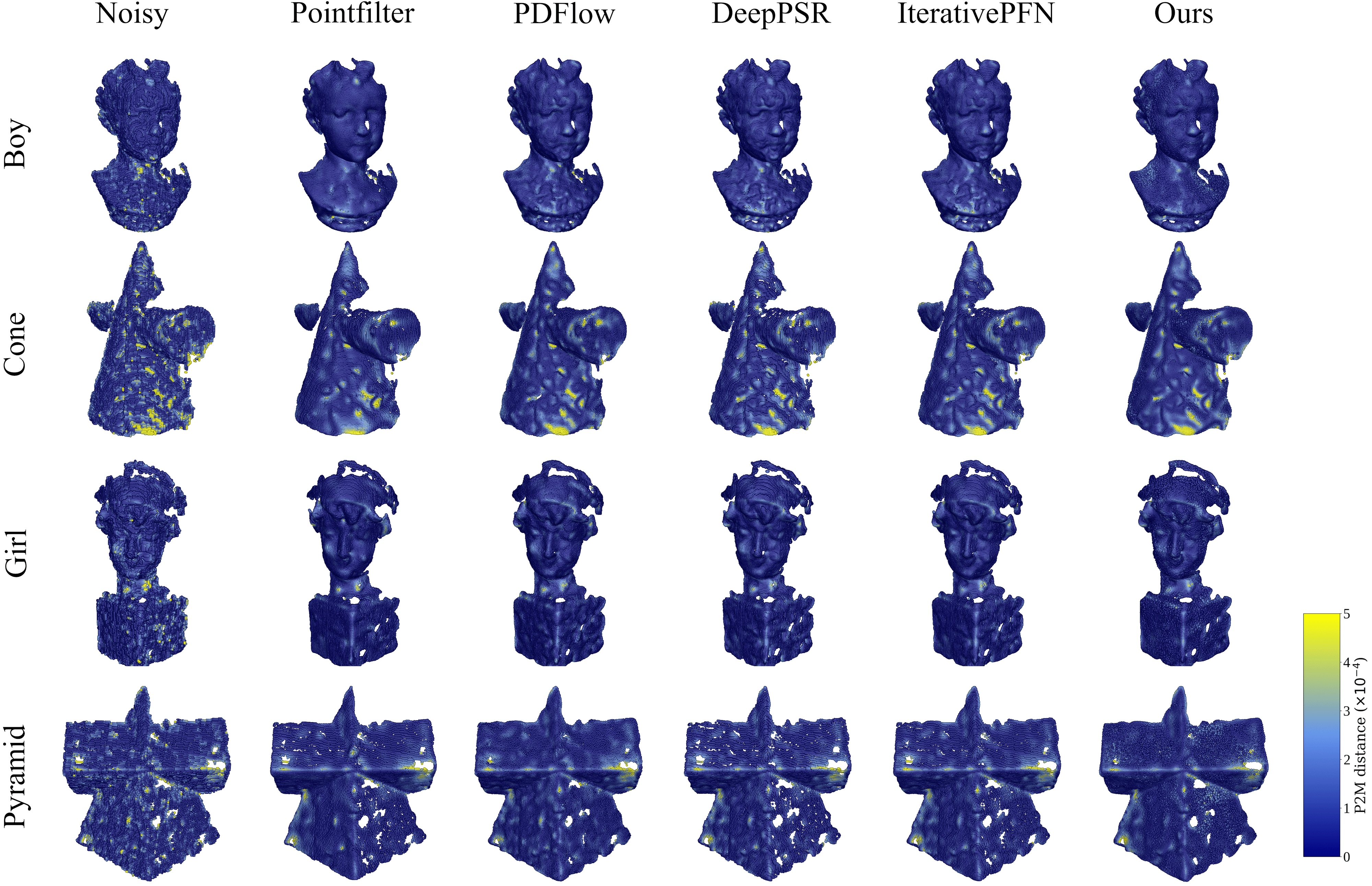}
\caption{Visual filtering results on 4 scans of the real-world Kinect data. We outperform other methods and recover better distributed points with fewer, and smaller, holes.}
\label{fig:supp-kinect-results}
\end{figure*}

\midsepremove
\begin{table*}[!tp]
\setlength{\tabcolsep}{5pt}
\centering
\begin{tabular}{c|l| cccccc|cccccc}
\toprule
\multicolumn{2}{c|}{Resolution} & \multicolumn{6}{c|}{10K (Sparse)} & \multicolumn{6}{c}{50K (Dense)} \\
\midrule
\multicolumn{2}{c|}{Noise} & \multicolumn{2}{c}{1\%} & \multicolumn{2}{c}{2\%} & \multicolumn{2}{c|}{3\%} & \multicolumn{2}{c}{1\%} & \multicolumn{2}{c}{2\%} & \multicolumn{2}{c}{3\%} \\
\midrule
\multicolumn{2}{c|}{Method} & CD & P2M & CD & P2M & CD & P2M & CD & P2M & CD & P2M & CD & P2M \\
\midrule
\parbox[t]{2mm}{\multirow{6}{*}{\rotatebox[origin=c]{90}{PUNet~\cite{Yu-PUNet}}}}
    & Score~\cite{Luo-Score-Based-Denoising}     & 2.479 & 0.463 & 3.701 & 1.098 & 4.764 & 1.984 & 0.711 & 0.149 & 1.319 & 0.593 & 2.087 & 1.162 \\
    & PointFilter~\cite{Zhang-Pointfilter}       & 2.399 & 0.433 & 3.529 & 0.888 & 5.240 & 2.001 & 0.757 & 0.185 & 0.964 & 0.301 & 1.918 & 0.963 \\
    & PDFlow~\cite{Mao-PDFlow}                   & 2.103 & 0.383 & 3.293 & 1.060 & 4.640 & 2.182 & 0.653 & 0.167 & 1.197 & 0.604 & 2.080 & 1.349 \\
    & DeepPSR~\cite{Chen-DeepPSR}                & 2.377 & 0.328 & 3.400 & 0.813 & \uline{4.179} & \uline{1.384} & 0.644 & \uline{0.075} & 1.036 & 0.345 & \bf 1.521 & \bf 0.692 \\
    & IterativePFN~\cite{Edirimuni-IterativePFN} & \uline{1.994} & \bf 0.212 & \uline{3.030} & \bf 0.567 & 4.611 & 1.678 & \uline{0.602} & \bf 0.061 & \uline{0.830} & \bf 0.205 & 2.425 & 1.377 \\
\cmidrule{2-14}
    & \bf Ours                                   & \bf 1.834 & \uline{0.238} & \bf 2.640 & \uline{0.617} & \bf 3.479 & \bf 1.292 & \bf 0.556 & 0.108 & \bf 0.798 & \uline{0.296} & \uline{1.662} & \uline{0.926} \\ 
\midrule
\parbox[t]{2mm}{\multirow{6}{*}{\rotatebox[origin=c]{90}{PCNet~\cite{Rakotosaona-PCN}}}}
    & Score~\cite{Luo-Score-Based-Denoising}     & 3.380 & 0.831 & 5.168 & 1.218 & 6.816 & 1.980 & 1.079 & 0.178 & 1.740 & 0.394 & 2.692 & 0.761 \\
    & PointFilter~\cite{Zhang-Pointfilter}       & 3.016 & 0.878 & 4.949 & 1.329 & 7.482 & 2.225 & 1.069 & 0.194 & 1.457 & 0.310 & 2.720 & \uline{0.691} \\
    & PDFlow~\cite{Mao-PDFlow}                   & 3.246 & \uline{0.607} & 4.722 & \uline{0.993} & 6.390 & \uline{1.738} & 0.988 & 0.159 & 1.734 & 0.478 & 2.828 & 0.811 \\
    & DeepPSR~\cite{Chen-DeepPSR}                & 3.146 & 0.993 & 4.943 & 1.342 & \uline{6.307} & 1.860 & 1.011 & 0.173 & 1.630 & 0.432 & \uline{2.332} & 0.724 \\
    & IterativePFN~\cite{Edirimuni-IterativePFN} & \bf 2.639 & 0.694 & \uline{4.492} & 1.036 & 6.534 & 1.910 & \uline{0.922} & \bf 0.142 & \uline{1.323} & \bf 0.272 & 3.046 & 0.987 \\
\cmidrule{2-14}
    & \bf Ours                                   & \uline{2.757} & \bf 0.543 & \bf 4.080 & \bf 0.830 & \bf 5.209 & \bf 1.329 & \bf 0.883 & \bf 0.148 & \bf 1.229 & \uline{0.290} & \bf 2.226 & \bf 0.654 \\
\bottomrule
\end{tabular}
\caption{Quantitative filtering results of recent methods and our method on the synthetic datasets with non-isotropic Gaussian noise. Our network is lightweight, with just $\sim 530K$ parameters (17\% of IterativePFN). CD and P2M values are multiplied by $10^4$. }
\label{tab:supp-pu-pc-cov-results}
\end{table*}
\midsepdefault

\midsepremove
\begin{table*}[!tp]
\setlength{\tabcolsep}{5pt}
\centering
\begin{tabular}{c|l| cccccc|cccccc}
\toprule
\multicolumn{2}{c|}{Resolution} & \multicolumn{6}{c|}{10K (Sparse)} & \multicolumn{6}{c}{50K (Dense)} \\
\midrule
\multicolumn{2}{c|}{Noise} & \multicolumn{2}{c}{1\%} & \multicolumn{2}{c}{2\%} & \multicolumn{2}{c|}{3\%} & \multicolumn{2}{c}{1\%} & \multicolumn{2}{c}{2\%} & \multicolumn{2}{c}{3\%} \\
\midrule
\multicolumn{2}{c|}{Method} & CD & P2M & CD & P2M & CD & P2M & CD & P2M & CD & P2M & CD & P2M \\
\midrule
\parbox[t]{2mm}{\multirow{6}{*}{\rotatebox[origin=c]{90}{PUNet~\cite{Yu-PUNet}}}}
    & Score~\cite{Luo-Score-Based-Denoising}     & 2.903 & 0.670 & 4.602 & 1.812 & 6.295 & 3.233 & 0.825 & 0.233 & 1.677 & 0.885 & 2.669 & \uline{1.645} \\
    & PointFilter~\cite{Zhang-Pointfilter}       & 2.773 & 0.571 & 4.250 & 1.366 & 7.668 & 3.905 & 0.827 & 0.230 & 1.244 & 0.491 & 2.876 & 1.800 \\
    & PDFlow~\cite{Mao-PDFlow}                   & 2.538 & 0.574 & 4.319 & 1.861 & 7.817 & 4.819 & 0.821 & 0.296 & 1.507 & 0.860 & 4.532 & 3.520 \\
    & DeepPSR~\cite{Chen-DeepPSR}                & 2.725 & 0.488 & 3.915 & 1.218 & \uline{4.982} & \bf 1.981 & 0.740 & 0.139 & 1.279 & 0.526 & \bf 1.835 & \bf 0.890 \\
    & IterativePFN~\cite{Edirimuni-IterativePFN} & 2.393 & \bf 0.315 & \uline{3.396} & \bf 0.806 & 6.489 & 3.131 & \uline{0.653} & \bf 0.089 & \uline{0.999} & \bf 0.317 & 3.670 & 2.411 \\
\cmidrule{2-14}
    & \bf Ours                                   & \bf 2.164 & \uline{0.338} & \bf 2.982 & \uline{0.872} & \bf 4.574 & \uline{2.163} & \bf 0.602 & \uline{0.133} & \bf 0.966 & \uline{0.412} & \uline{2.536} & 1.676 \\ 
\midrule
\parbox[t]{2mm}{\multirow{6}{*}{\rotatebox[origin=c]{90}{PCNet~\cite{Rakotosaona-PCN}}}}
    & Score~\cite{Luo-Score-Based-Denoising}     & 3.968 & 0.954 & 6.193 & 1.650 & 8.395 & 2.618 & 1.204 & 0.227 & 2.058 & 0.509 & 3.379 & 1.013 \\
    & PointFilter~\cite{Zhang-Pointfilter}       & 3.539 & 0.985 & 5.732 & 1.589 & 9.821 & 3.273 & 1.155 & 0.227 & 1.723 & 0.381 & 3.536 & 1.099 \\
    & PDFlow~\cite{Mao-PDFlow}                   & 3.760 & \uline{0.713} & 5.778 & 1.404 & 9.245 & 2.869 & 1.166 & 0.214 & 2.107 & 0.544 & 4.831 & 1.453 \\
    & DeepPSR~\cite{Chen-DeepPSR}                & 3.687 & 1.133 & 5.452 & 1.573 & \uline{7.142} & \uline{2.166} & 1.121 & 0.226 & 1.830 & 0.484 & \bf 2.574 & \bf 0.740 \\
    & IterativePFN~\cite{Edirimuni-IterativePFN} & \uline{3.189} & 0.801 & \uline{4.894} & \uline{1.165} & 8.173 & 2.517 & \uline{0.993} & \uline{0.172} & \uline{1.477} & \bf 0.288 & 3.799 & 1.138 \\
\cmidrule{2-14}
    & \bf Ours                                   & \bf 3.156 & \bf 0.602 & \bf 4.468 & \bf 0.940 & \bf 6.099 & \bf 1.548 & \bf 0.936 & \bf 0.161 & \bf 1.399 & \uline{0.332} & \uline{2.916} & \uline{0.892} \\
\bottomrule
\end{tabular}
\caption{Quantitative filtering results of recent methods and our method on the synthetic datasets with Laplace noise. Our network is lightweight, with just $\sim 530K$ parameters (17\% of IterativePFN). CD and P2M values are multiplied by $10^4$. }
\label{tab:supp-pu-pc-laplace-results}
\end{table*}
\midsepdefault

\subsection{Additional visual results on real-world scanned data}
Fig.~\ref{fig:supp-rm-results} illustrates filtering results on Scene 3 and Scene 4 of the Paris-Rue-Madame dataset. We observe from the close-ups in Scene 3 and 4 that PDFlow and DeepPSR leave behind high amounts of noise near the building windows. By comparison, IterativePFN fairs better but is not able to completely filter these noisy artifacts. StraightPCF is able to recover cleaner, smoother surfaces and removes a higher proportion of noise. This is evident in the close-ups of the windows and the close-up of the car roof. Further evaluation on real-world outdoor scenes is provided in Fig.~\ref{fig:supp-kitti-results} which illustrates filtering results on 4 scenes of the Kitti-360 dataset~\cite{Liao-Kitti360}. This dataset contains point clouds at a high sparsity setting and high noise level. While other methods leave behind noisy remnants, StraightPCF provides smoothly filtered surfaces. Moreover, the points are better distributed, as compared with DeepPSR which leaves behind clustering artifacts. Next, Fig.~\ref{fig:supp-kinect-results} demonstrates visual filtering results on the Kinect data. 
Methods such as PDFlow, DeepPSR and IterativePFN perform poorly on scans such as Boy and Pyramid, in comparison to StraightPCF. Furthermore, Pointfilter leaves behind a large number of holes on the Pyramid scan whereas StraightPCF recovers a cleaner filtered version of the Pyramid 
with fewer, and smaller, holes.

\midsepremove
\begin{table*}[!tp]
\setlength{\tabcolsep}{5pt}
\centering
\begin{tabular}{c|l| cccccc|cccccc}
\toprule
\multicolumn{2}{c|}{Resolution} & \multicolumn{6}{c|}{10K (Sparse)} & \multicolumn{6}{c}{50K (Dense)} \\
\midrule
\multicolumn{2}{c|}{Noise} & \multicolumn{2}{c}{1\%} & \multicolumn{2}{c}{2\%} & \multicolumn{2}{c|}{3\%} & \multicolumn{2}{c}{1\%} & \multicolumn{2}{c}{2\%} & \multicolumn{2}{c}{3\%} \\
\midrule
\multicolumn{2}{c|}{Method} & CD & P2M & CD & P2M & CD & P2M & CD & P2M & CD & P2M & CD & P2M \\
\midrule
\parbox[t]{2mm}{\multirow{6}{*}{\rotatebox[origin=c]{90}{PUNet~\cite{Yu-PUNet}}}}
    & Score~\cite{Luo-Score-Based-Denoising}     & 1.274 & 0.249 & 2.467 & 0.414 & 3.366 & 0.977 & 0.505 & 0.046 & 0.691 & 0.129 & 0.900 & 0.288 \\
    & PointFilter~\cite{Zhang-Pointfilter}       & 1.139 & 0.287 & 2.447 & 0.407 & 3.031 & 0.574 & 0.631 & 0.148 & 0.742 & 0.171 & 0.798 & \uline{0.188} \\
    & PDFlow~\cite{Mao-PDFlow}                   & 0.874 & 0.178 & 2.026 & 0.332 & 2.660 & 0.637 & 0.456 & 0.057 & 0.854 & 0.326 & 0.953 & 0.400 \\
    & DeepPSR~\cite{Chen-DeepPSR} & 1.226 & 0.190 & 2.416 & 0.299 & 2.969 & 0.472 & 0.497 & \uline{0.027} & 0.638 & \uline{0.064} & 0.928 & 0.245 \\
    & IterativePFN~\cite{Edirimuni-IterativePFN} & 0.645 & \bf 0.088 & \uline{2.009} & \bf 0.191 & \uline{2.684} & \bf 0.339 & \uline{0.443} & \bf 0.014 & \uline{0.599} & 0.054 & \uline{0.694} & \bf 0.110 \\
\cmidrule{2-14}
    & \bf Ours                                   & \bf \uline{0.624} & \uline{0.097} & \bf 1.834 & \uline{0.252} & \bf 2.365 & \uline{0.427} & \bf 0.418 & 0.050 & \bf 0.557 & 0.130 & \bf 0.683 & 0.228 \\ 
\midrule
\parbox[t]{2mm}{\multirow{6}{*}{\rotatebox[origin=c]{90}{PCNet~\cite{Rakotosaona-PCN}}}}
    & Score~\cite{Luo-Score-Based-Denoising}     & 1.789 & 0.692 & 3.232 & 0.795 & 4.767 & 1.258 & 0.714 & 0.104 & 1.044 & 0.164 & 1.342 & 0.308 \\
    & PointFilter~\cite{Zhang-Pointfilter}       & 1.414 & 0.792 & 2.926 & 0.853 & 4.032 & 0.999 & 0.774 & 0.155 & 1.060 & 0.181 & 1.182 & \uline{0.210} \\
    & PDFlow~\cite{Mao-PDFlow}                   & 2.038 & \uline{0.506} & 3.108 & \uline{0.582} & 3.954 & \uline{0.788} & 0.727 & \uline{0.096} & 1.245 & 0.329 & 1.374 & 0.361 \\
    & DeepPSR~\cite{Chen-DeepPSR}                & 1.610 & 0.954 & 3.065 & 1.014 & 4.146 & 1.129 & \uline{0.661} & 0.110 & 0.991 & \uline{0.157} & 1.306 & 0.270 \\
    & IterativePFN~\cite{Edirimuni-IterativePFN} & \bf 0.989 & 0.598 & \bf 2.495 & 0.668 & \uline{3.713} & 0.805 & \bf 0.578 & \bf 0.089 & \uline{0.918} & \bf 0.134 & \uline{1.074} & \bf 0.186 \\
\cmidrule{2-14}
    & \bf Ours                                   & \uline{1.399} & \bf 0.477 & \uline{2.917} & \bf 0.581 & \bf 3.677 & \bf 0.700 & 0.678 & 0.103 & \bf 0.906 & 0.175 & \bf 1.058 & 0.240 \\
\bottomrule
\end{tabular}
\caption{Quantitative filtering results of recent methods and our method on} the synthetic 
datasets with uniform noise. 
Our network is lightweight, with just $\sim 530K$ parameters (17\% of IterativePFN). CD and P2M values are multiplied by $10^4$.
\label{tab:supp-pu-pc-uniform-results}
\end{table*}
\midsepdefault

\section{Further Evaluation on PUNet and PCNet Data Under Different Noise Patterns}
\begin{figure*}[!tp]
\centering
\includegraphics[width=0.97\linewidth]{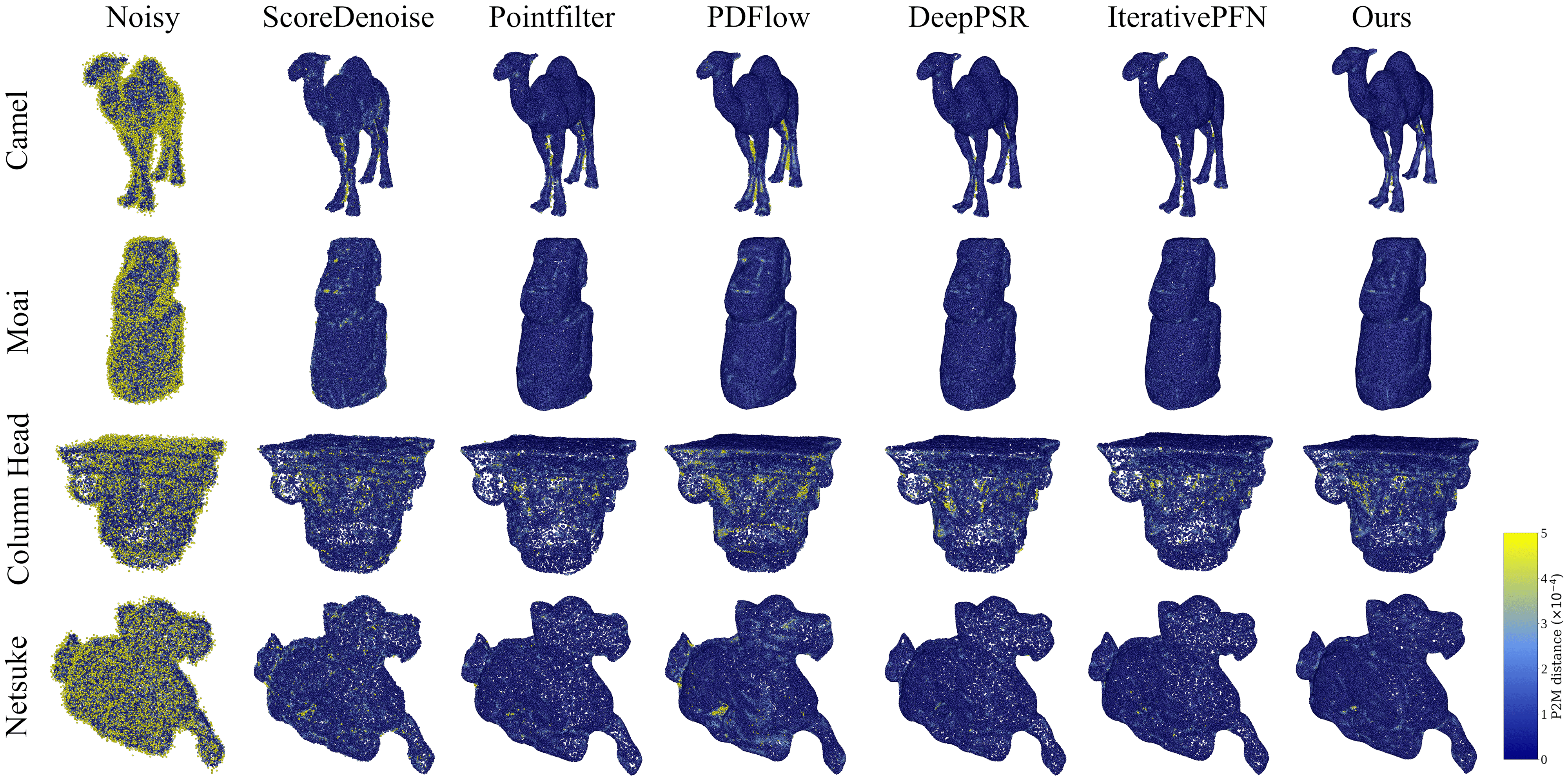}
\caption{Visual results for 50K resolution shapes with non-isotropic Gaussian noise and noise scale $s=2\%$ of the bounding sphere radius. }
\label{fig:supp-synthetic-cov-results}
\end{figure*}

Similar to the analysis of ScoreDenoise~\cite{Luo-Score-Based-Denoising}, DeepPSR~\cite{Chen-DeepPSR}, PDFlow~\cite{Mao-PDFlow} and IterativePFN~\cite{Edirimuni-IterativePFN}, we look at quantitative and visual results on synthetic data under different noise patterns. More specifically, we look at noise patterns that have the following distributions:

\textbf{1) Non-isotropic Gaussian distribution} where the covariance matrix is given by:
\begin{equation}
    \Sigma = s^2\times \begin{bmatrix}
                        1 & -1/2 & -1/4 \\
                        -1/2 & 1 & -1/4 \\
                        -1/4 & -1/4 & 1
                        \end{bmatrix}
\end{equation}
The noise scale $s$ is set to $1\%$, $2\%$ and $3\%$ of the bounding sphere's radius. In contrast to the synthetic data presented in the main paper, which contain isotropic Gaussian noise, here we look at Gaussian noise that is anisotropically distributed. Table~\ref{tab:supp-pu-pc-cov-results} and \cref{fig:supp-synthetic-cov-results} provide quantitative and visual results on this noise pattern. Our method outperforms others across most resolution and noise settings on both the PUNet and PCNet datasets.

\begin{figure*}[!tp]
\centering
\includegraphics[width=0.97\linewidth]{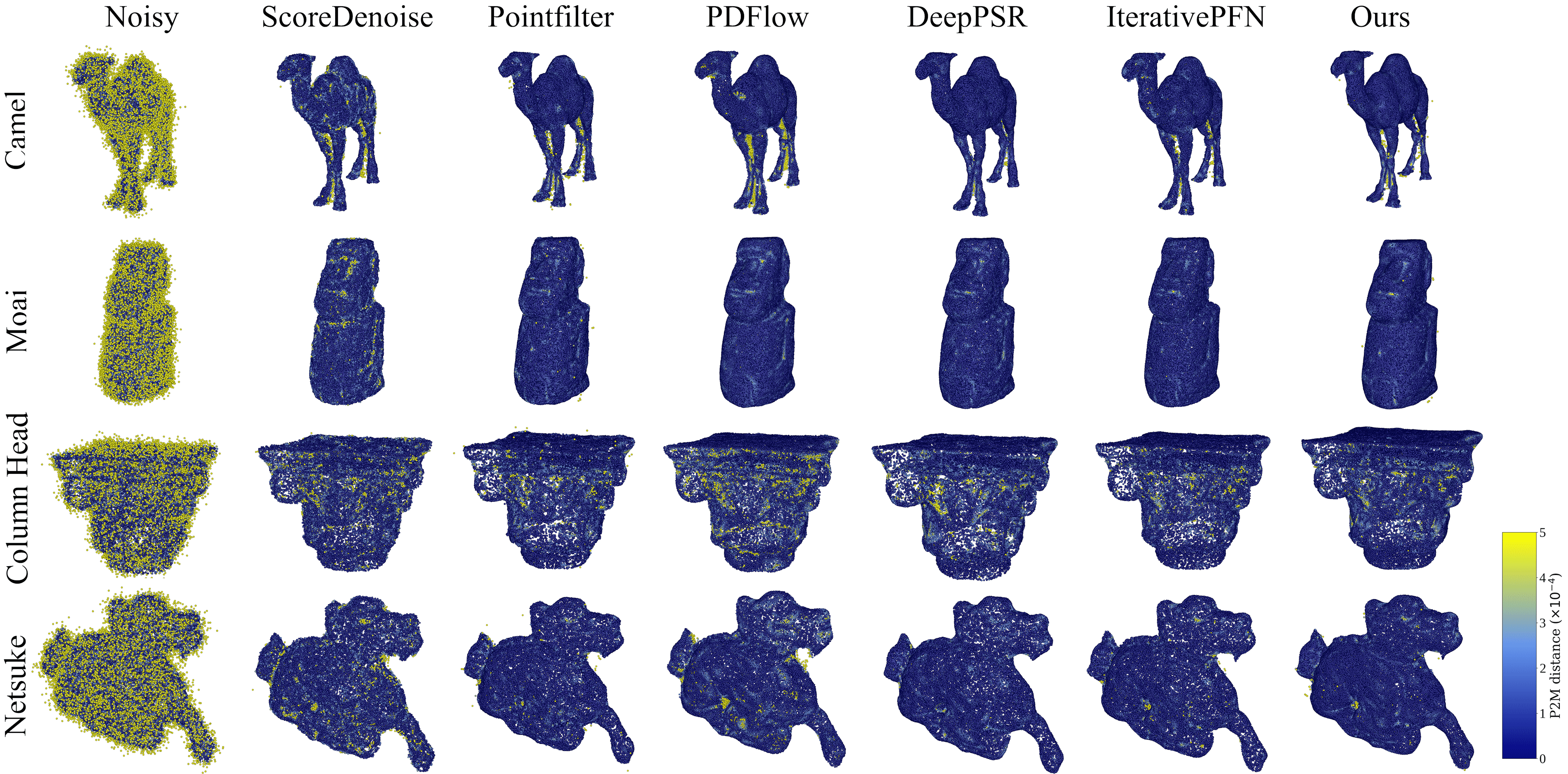}
\caption{Visual results for 50K resolution shapes with Laplace noise and noise scale $s=2\%$ of the bounding sphere radius. }
\label{fig:supp-synthetic-laplace-results}
\end{figure*}

\textbf{2) Laplace distribution} where the noise scale $s$ is set to $1\%$, $2\%$ and $3\%$ of the bounding sphere's radius. Table~\ref{tab:supp-pu-pc-laplace-results} and \cref{fig:supp-synthetic-laplace-results} provide quantitative and visual results on this noise pattern. We note that the noise intensity for this noise pattern is generally higher than the case of Gaussian noise. However, our method satisfactorily filters synthetic data across both PUNet and PCNet datasets. IterativePFN and DeepPSR obtain competitive P2M results, yet induce clustering which can be seen from the formation of small holes on the Camel and Netsuke shapes of \cref{fig:supp-synthetic-laplace-results}.  We also observe that while our StraightPCF method interpolates between Gaussian high noise variant patches and underlying clean patches during training time, the results on the Laplace distributed synthetic data demonstrates the high generalizability of our method. 

\begin{figure*}[!tp]
\centering
\includegraphics[width=0.97\linewidth]{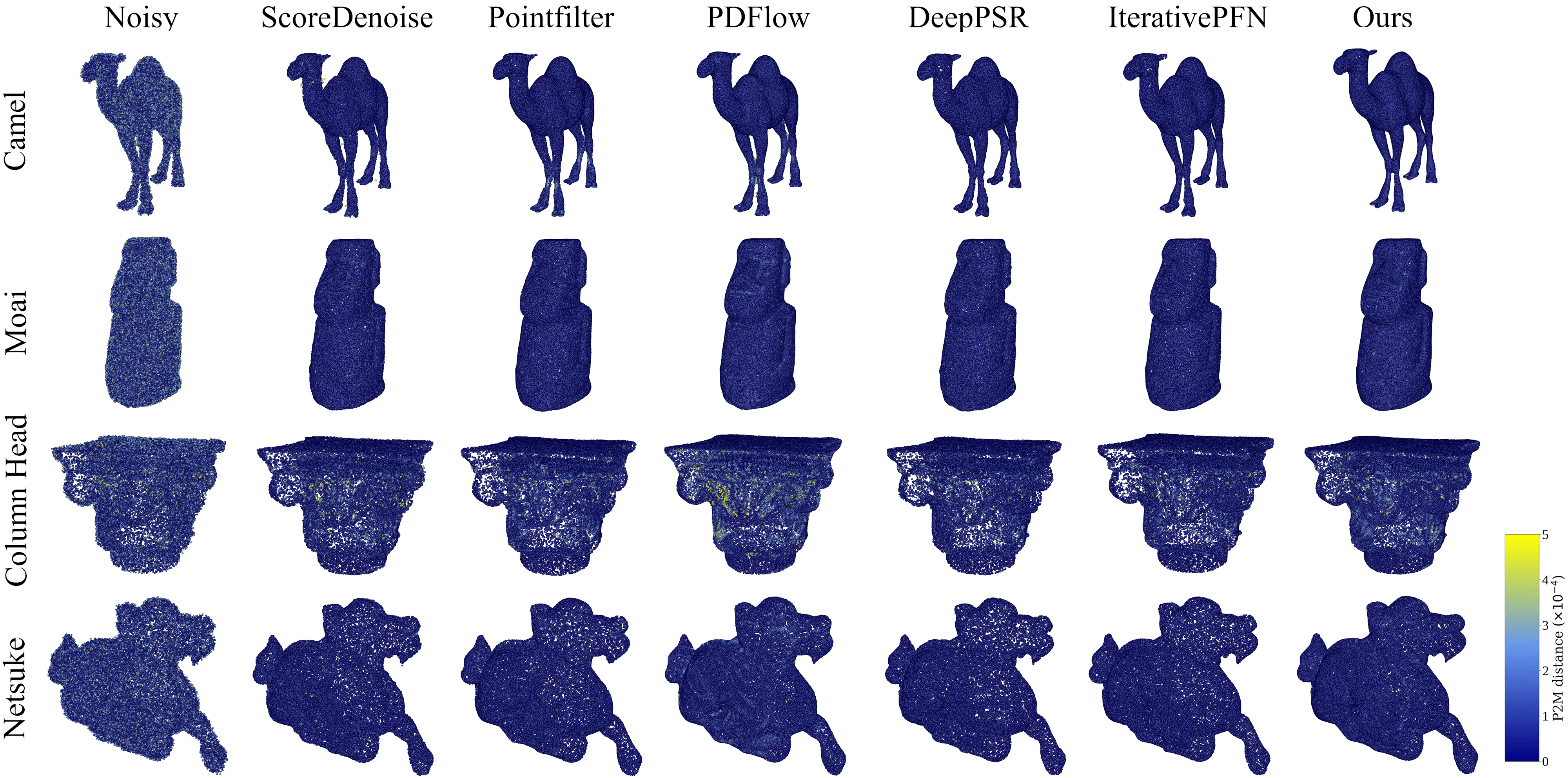}
\caption{Visual results for 50K resolution shapes with uniformly distributed noise and noise scale $s=2\%$ of the bounding sphere radius. }
\label{fig:supp-synthetic-uniform-results}
\end{figure*}

\textbf{3) Uniform distribution} of noise within a sphere of radius $s$. The probability to sample noise at a position $\pmb{x}$ within the sphere is given by,
\begin{equation}
    p(\pmb{x}; s) = \begin{cases} \frac{3}{4\pi s^3}, &\norm{\pmb{x}}_2\leq s, \\
    0, &\text{Otherwise}
    \end{cases}
\end{equation}
where the noise scale $s$ is set to $1\%$, $2\%$ and $3\%$ of the bounding sphere's radius. This noise distribution is not uni-modal unlike the previous distributions and generally has a lower noise intensity than that of Gaussian noise. Table~\ref{tab:supp-pu-pc-uniform-results} and \cref{fig:supp-synthetic-uniform-results} provide quantitative and visual results on this noise pattern. Overall, StraightPCF consistently outperforms others on the Chamfer Distance metric, indicating its ability to recover a distribution of points closer to that of the clean point cloud. Furthermore, analysis of the visual results again reinforces the conclusion that while some methods such as IterativePFN may yield lower P2M errors, they cause points to cluster and leave behind small holes.

\section{Comparison of Test Times for Different Methods}

\midsepremove
\begin{table}[!tp]
\centering
\begin{tabular}{l|ll|ll}
\toprule
Method & Time (s) \\
\midrule
PCN & 186.7 \\
ScoreDenoise & 15.8 \\
Pointfilter & 100.8 \\
PDFlow & 53.8 \\
DeepPSR & 8.99 \\
IterativePFN & 19.7 \\
\midrule
Ours & 18.2 \\
\bottomrule
\end{tabular}
\caption{Runtimes of state-of-the-art methods on point clouds with 50K points and 2\% Gaussian noise, from the PUNet dataset.}
\label{tab:runtimes} 
\end{table}
\midsepdefault

\label{sec:comp-test-times}
Table~\ref{tab:runtimes} provides runtimes for StraightPCF, and other state-of-the-art methods, on synthetic point clouds at 50K resolution and $\sigma=2\%$. We obtained the runtime for each method by filtering 3 point clouds at 50K resolution and $\sigma=2\%$, from the PUNet dataset, and taking the mean of the total runtime. In general, DeepPSR is the fastest method but its overall performance across synthetic and scanned data is sub-optimal. To improve performance results, DeepPSR would potentially need to filter point clouds for additional iterations, leading to higher runtimes. The same is true for ScoreDenoise. However, we obtain state-of-the-art performance with highly competitive runtimes. As mentioned previously, StraightPCF is much more lightweight than IterativePFN, its closest competitor in terms of performance. It has only $\sim530K$ parameters compared to IterativePFN's $\sim3.2M$ parameters. 

\midsepremove
\begin{table}[!tp]
\centering
\setlength\tabcolsep{4pt} 
\begin{tabular}{c|ll|ll|ll}
\toprule
\multirow{3}{*}{Ablation: PUNet} & \multicolumn{6}{c}{10K points} \\
\cmidrule{2-7}
& \multicolumn{2}{c|}{1\% noise} & \multicolumn{2}{c|}{2\% noise} & \multicolumn{2}{c}{3\% noise} \\
\cmidrule{2-7}
& CD & P2M & CD & P2M & CD & P2M \\ 
\midrule
\textbf{\textit{V5})} \textbf{CVM w/ DM}   & \uline{1.87} & \bf 0.24 & \uline{2.64} & \uline{0.60} & \uline{3.29} & \uline{1.13}  \\
V6) 3VM w/o DM  & 1.89 & 0.27 & 2.81 & 0.76 & 3.43 & 1.25 \\
V7) 3VM w/ DM   & \uline{1.87} & \bf 0.24 & 2.66 & 0.61 & \bf 3.23 & \bf 1.09 \\
V8) 4VM w/o DM  & 1.90 & 0.28 & 2.87 & 0.83 & 3.52 & 1.34 \\
V9) 4VM w/ DM   & \bf 1.86 & \bf 0.24 & \bf 2.62 & \bf 0.59 & 3.38 & 1.20 \\
\bottomrule
\end{tabular}
\caption{Ablation results for higher VelocityModule (VM) couplings $K$, with and without the DistanceModule (DM). CD and P2M distances are multiplied by $10^4$.}
\label{tab:supp-ablation-higher-couplings} 
\end{table}
\midsepdefault

\midsepremove
\begin{table}[!tp]
\centering
\setlength\tabcolsep{4pt} 
\begin{tabular}{c|ll|ll|ll}
\toprule
\multirow{3}{*}{Ablation: PUNet} & \multicolumn{6}{c}{10K points} \\
\cmidrule{2-7}
& \multicolumn{2}{c|}{1\% noise} & \multicolumn{2}{c|}{2\% noise} & \multicolumn{2}{c}{3\% noise} \\
\cmidrule{2-7}
& CD & P2M & CD & P2M & CD & P2M \\ 
\midrule
$N=1$ & 2.03 & 0.30 & 2.84 & 0.76 &  3.44 & 1.26 \\
\textbf{$N=3$}  & \uline{1.87} & \bf 0.24 & \uline{2.64} & \uline{0.60} & \bf 3.29 & \bf 1.13 \\
$N=5$  & \bf 1.86 & \bf 0.24 & \bf 2.62 & \bf 0.59 & \uline{3.33} & \uline{1.15} \\
\bottomrule
\end{tabular}
\caption{Ablation on different numbers of Euler steps $N$. Based on the above results, we see that $N=3$ is the optimal number of Euler steps as it provides a good balance between performance and efficiency. CD and P2M distances are multiplied by $10^4$.}
\label{tab:supp-ablation-euler-steps} 
\end{table}
\midsepdefault

\begin{figure}[!tp]
\centering
\includegraphics[width=0.97\linewidth]{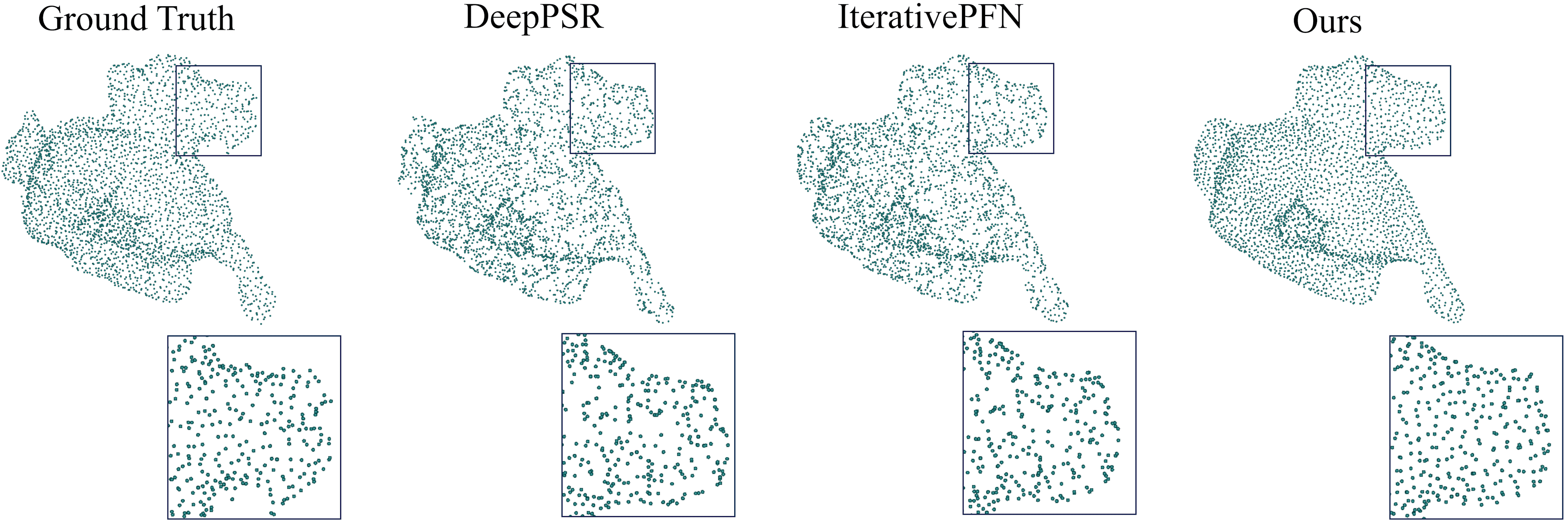}
\caption{Visual results for the 3K resolution Netsuke shape with Gaussian noise and $\sigma=3\%$ of the bounding sphere radius. Filtering sparse point clouds at high noise is a challenge for StraightPCF. However, it recovers a better distribution of points, as compared with other state-of-the-art methods.}
\label{fig:supp-limitations}
\end{figure}

\section{Further Ablation Studies}
\label{sec:further-ablations}

Tables~\ref{tab:supp-ablation-higher-couplings} and~\ref{tab:supp-ablation-euler-steps} provide ablation results on higher VelocityModule (VM) couplings $K$, with and without the DistanceModule, and the impact of tuning the number of Euler steps $N$, respectively. We see that the optimal architecture is V5 as it only contains $K=2$ VMs and is smaller than the other variants, leading to a faster runtime. Generally, it also provide either the best or second best results on the ablation set. The results of Tables~\ref{tab:supp-ablation-higher-couplings} further reinforces the importance of the DistanceModule to ensure convergence at the clean surface, especially as noise increases. Moreover, Table~\ref{tab:supp-ablation-euler-steps} demonstrates the importance of Euler steps $N$. For $N=1$, we obtain sub-optimal results. For $N=3$, we strike a balance between performance and runtime efficiency as runtime increases as $N$ increases. Also, we see only a marginal improvement in performance for $N>3$.

\section{Discussion of Limitations}
\label{sec:limitations-discussion}
Finally, \cref{fig:supp-limitations} illustrates the visual filtering results for the Netsuke shape with 3K resolution and $\sigma=3\%$. We see that at this high sparsity setting, it is difficult to recover high levels of geometric detail. However, our method performs better than other state-of-the-art methods and recovers a nicer distribution of points.